%% file: example_paper.tex
\documentclass{article}

\usepackage{microtype}
\usepackage{graphicx}
\usepackage{booktabs} 
\usepackage{caption}
\usepackage{subcaption}

\usepackage{hyperref}
\usepackage[accepted]{mlsys2025}
    
\usepackage{tikz}
\usepackage{amsmath}
\usepackage{amssymb}
\usepackage{booktabs}
\usepackage{ragged2e}
\usepackage{array}
\usepackage{multicol}
\usepackage{pifont}
\usepackage{xspace}
\usepackage{multirow}
\usepackage{graphicx}
\usepackage{enumitem}
\usepackage{balance}
\usepackage{xcolor}
\usepackage{colortbl}

\usepackage{algorithm}
\usepackage{amsmath}        
\usepackage{xurl} 
\usepackage{listings}
\renewcommand{\algorithmicrequire}{\textbf{Input:}}
\renewcommand{\algorithmicensure}{\textbf{Output:}}
\definecolor{ForestGreen}{RGB}{34,139,34}
\providecommand{\algcomment}[1]{\hspace{1em}\textcolor{ForestGreen}{// #1}}

\newcommand{\niparagraph}[1]{\vspace{1pt}\noindent\textbf{#1}}

\newcommand{\nest}{\textsc{Nest}\xspace}
\newcommand{\subgraph}{\textsc{Sub-Graph}\xspace}
\newcommand{\graphglobal}{\textsc{Graph-Global}\xspace}
\newcommand{\cmark}{\ding{52}}
\newcommand{\xmark}{\ding{56}}
\definecolor{mygreen}{cmyk}{1, 0, 1, 0} 

\mlsystitlerunning{\nest}
\input{badges}
\begin{document}

\twocolumn[
\mlsystitle{NEST: Network- and Memory-Aware Device Placement \\ For Distributed Deep Learning}




 \begin{mlsysauthorlist}
\mlsysauthor{Irene Wang}{gt}
\mlsysauthor{Vishnu Varma Venkata}{gt}
\mlsysauthor{Arvind Krishnamurthy}{uw}
\mlsysauthor{Divya Mahajan}{gt}
\end{mlsysauthorlist}

\mlsysaffiliation{gt}{Georgia Institute of Technology, Atlanta, GA, USA}
\mlsysaffiliation{uw}{University of Washington, Seattle, WA, USA}

\mlsyscorrespondingauthor{Irene Wang}{irene.wang@gatech.edu}

\mlsyskeywords{Distributed ML Training}

\vspace{3ex}

\input{body/abstract}

]

\printAffiliationsAndNotice{}
\input{body/intro}
\input{body/background}

\input{body/design}

\input{body/evaluation}

\input{body/relatedWorks}

\input{body/conclusion}

\newpage
\balance
\bibliography{example_paper}
\bibliographystyle{mlsys2025}

\clearpage
\input{body/appendix}


\end{document}

%% file: body/abstract.tex
\begin{abstract}

The growing scale of deep learning demands distributed training frameworks that jointly reason about parallelism, memory, and network topology. Prior works often rely on heuristic or topology-agnostic search~\cite{pip, alpa, phaze}, handling communication and memory separately. Without per-device memory awareness, these methods typically ensure feasibility post hoc by sharding parameters and activations across many devices, increasing synchronization, inflating communication, and underutilizing compute—limiting scalability and efficiency on real datacenter networks.
We present \nest, a network-, compute-, and memory-aware device placement framework that unifies model parallelism, topology modeling, and memory feasibility via structured dynamic programming (DP). \nest{}’s DP operates on operator graphs annotated with intra-layer parallelism configurations (tensor, expert, sequence, context), explicit allreduce latencies across hierarchical or arbitrary networks, and memory/compute profiles. By composing these with pipeline, data, and ZeRO partitioning, \nest defines a principled search space for hybrid strategies while jointly optimizing co-location, network latency, and memory feasibility.
Evaluations across diverse hardware and networks show \nest achieves up to 2.43$\times$ higher throughput, better memory efficiency, and improved scalability over state-of-the-art baselines, providing a foundation for co-designing parallelization strategies and datacenter interconnects for next-generation AI infrastructure. 
The source code of \nest is available at: \href{https://github.com/scai-tech/Nest}{https://github.com/scai-tech/Nest}.

\end{abstract}

%% file: body/intro.tex
\section{Introduction}

\begin{figure*}[t]
    \centering
    \includegraphics[width=1\linewidth]{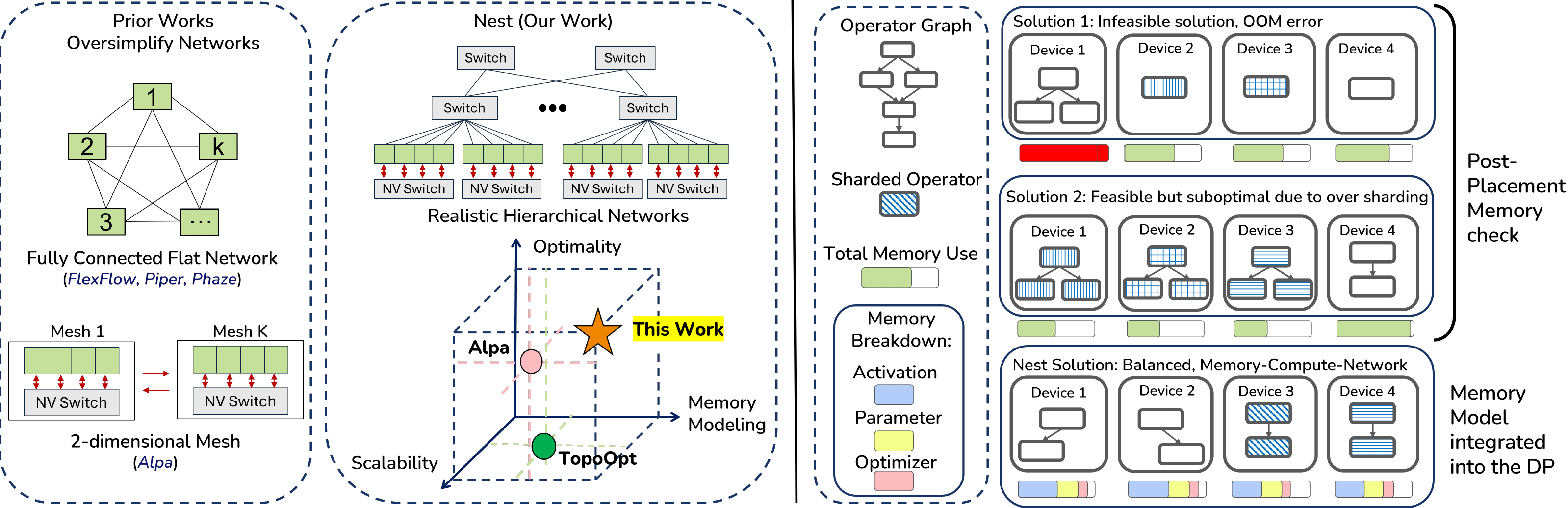}
    \vspace{-2ex}
    \caption{(Left) \nest compared to prior works across network modeling, optimality, scalability, and memory modeling axes. (Right) Comparison of placement with and without integrated memory modeling. Sharded operators are shown through patterns.}
    \label{fig:intro}
    \vspace{-2ex}
\end{figure*}


The rapid growth of deep learning in both parameter count and model complexity has made efficient distributed training a key bottleneck in scaling modern AI systems~\cite{megatronlm, nemo, deepspeed}. Models like GPT-3 and Llama3 are trained on clusters with over 10,000 GPUs~\cite{gpt3_training, llama3_training}, where efficiency depends on how models are partitioned across devices and how these partitions interact with the physical interconnect. Even with abundant compute, inefficient placement can make the network the performance bottleneck, increasing training time and reducing infrastructure utilization.

To distribute training, we leverage multiple parallelization strategies, tensor, pipeline, data, expert, and optimizer-level schemes like ZeRO~\cite{gpipe, pipedream, dataparallel, gshard, zero}. Each of these introduces distinct communication patterns and memory trade-offs, and real-world training systems rely on complex hybrid combinations to meet scaling and efficiency targets. 
Performance thus depends critically on datacenter interconnects, which handle collectives such as all-reduce for gradients and reduce-scatter/all-gather for activations or expert exchanges~\cite{ccoverlap-ispass}. Yet, most existing device placement frameworks~\cite{piper, alpa, pip, phaze, phaze-patent} assume uniform or flat networks, overlooking the heterogeneous, hierarchical, and often oversubscribed nature of real-world topologies. For example, NVIDIA’s DGX SuperPOD links GPUs via NVSwitch within nodes and InfiniBand across nodes~\cite{superpod}; Microsoft’s MAIA clusters use hierarchical RDMA networks with bandwidth-aware routing~\cite{msftmaia}; and Google’s TPU v4 Pods employ torus topologies with non-uniform communication costs~\cite{tpuv4_isca}. These variations in latency and bandwidth can significantly impact performance-critical collectives, often making network interconnects a major bottleneck in distributed training~\cite{microChar}.


Recent works such as Alpa and TopoOpt~\cite{alpa, topoopt} introduce topology-aware device placement but remain limited in scalability and generality. TopoOpt uses Monte Carlo-based random search, offering no optimality guarantees and scaling poorly as the search space grows. Alpa assumes a simplified 2D mesh and validates communication-driven sharding only after placement, causing over-sharding, excessive communication, and underutilized compute—limiting scalability beyond ~64 GPUs or heterogeneous clusters.
As shown in Figure~\ref{fig:intro}, existing frameworks do not jointly handle model parallelism, memory constraints, and realistic network topology while ensuring scalable, near-optimal search. Bridging this gap is essential for improving training throughput, cost efficiency, and guiding co-design of model architectures and datacenter interconnects. A principled, topology-aware placement framework that integrates computation, communication, and memory can expose how network capabilities and training strategies should co-evolve for maximal efficiency.

As such, in this work, we introduce \nest, a principled, extensible, and compute–memory–network aware device placement framework that unifies model and network optimization through a novel dynamic programming formulation. \nest\!’s novel dynamic programming optimizer operates over the operator graph augmented with multiple tensor, sequence, expert, and context parallel configurations, precomputed communication latencies across network levels, and profiled compute and memory statistics. It inherently accounts for ZeRO for memory savings, pipeline parallelism for layer splitting, and data parallelism for scalability. This forms the combinatorial search space for the DP optimizer, enabling principled exploration of hybrid parallelism strategies under realistic network and memory constraints.


A key insight of \nest is its incremental handling of communication and memory costs. Since placement is solved backward, a layer’s communication depends on unknown downstream placements. \nest addresses this by precomputing latencies between abstract network levels (e.g., intra-node, intra-rack, and other topologies) and evaluating transitions across multiple candidate downstream placements, preserving both network accuracy and provable optimality.

To further enhance efficiency and extensibility, \nest integrates a detailed memory model that tracks activations, gradients, parameters, and optimizer states during optimization, rather than relying on post-placement checks. This allows adaptive application of ZeRO stages to unlock previously infeasible configurations. Parallelism strategies are categorized along orthogonal dimensions, enabling systematic composition of hybrid strategies without exploding the search space. Finally, \nest\!’s flexible network interface supports realistic datacenter topologies—including oversubscribed trees, fat-trees, spine-leaf fabrics, and torus networks—facilitating rigorous evaluation and co-design of model placement and datacenter infrastructure.

As such, the main contributions of \nest~are:
\vspace{-1ex}
\begin{enumerate}[leftmargin=*, itemsep=1pt, parsep=0pt, topsep=1pt, partopsep=1pt]
\item \textbf{Efficient Network-aware dynamic programming algorithm} with provable optimality guarantees for hierarchical, heterogeneous interconnects.
\item \textbf{Unified support for diverse parallelism strategies} (tensor, pipeline, data, expert, sequence, context, ZeRO) with structured composition for tractable large-scale search.
\item \textbf{Comprehensive network and memory modeling} that reflects production-scale topologies and enables systematic exploration of placement–infrastructure trade-offs.
\end{enumerate}
\vspace{-1ex}

We evaluate \nest on large-scale training workloads with billions of parameters in realistic datacenter settings. \nest consistently outperforms manual, Monte Carlo–based, and state-of-the-art dynamic programming baselines, achieving up to 2.43$\times$ higher throughput and sustaining performance beyond 1,000 GPUs. It also finds optimal placements under memory constraints by adaptively leveraging ZeRO stages.
%
%

\begin{table*}[ht]
\centering
\vspace{-2ex}
\caption{Comparison of prior network- and memory- aware device placement works.}
\scriptsize
\renewcommand{\arraystretch}{0.5}
\begin{tabular}{
    >{\RaggedRight\hsize=.11\hsize\linewidth=\hsize}m{\linewidth} 
    >{\RaggedRight\hsize=.19\hsize\linewidth=\hsize}m{\linewidth} 
    >{\RaggedRight\hsize=.19\hsize\linewidth=\hsize}m{\linewidth} 
    >{\RaggedRight\hsize=.2\hsize\linewidth=\hsize}m{\linewidth} 
    >{\columncolor{cyan!5}\RaggedRight\hsize=.19\hsize\linewidth=\hsize}m{\linewidth}
}
\toprule
\textbf{Feature} & \textbf{TopoOpt}\newline\cite{topoopt} & \textbf{Alpa}\newline\cite{alpa} & \textbf{Mist}\newline\cite{mist} & \textbf{\nest~(Ours)} \\
\midrule
\textbf{Parallelism Strategies} 
& Data, Pipeline, Operator  
& Pipeline, Intra-operator \mbox{(including Data)}
& Data, Pipeline, Tensor, ZeRO
& Data, Pipeline, Tensor, ZeRO, Expert, Sequence, Context \\
& & & & \cellcolor{cyan!5} \\
\textbf{Algorithm} 
& MCMC (random search) 
& DP + ILP (joint search)
& Hierarchical MILP + brute-force enumeration
& Dynamic Programming \\
& & & & \cellcolor{cyan!5} \\
\textbf{Network\newline Awareness \newline} 
& \textcolor{green}{\cmark} TopoOpt Specific
& \textcolor{orange}{\protect\ding{110}} 2D mesh only
& \textcolor{red}{\xmark} Profiles interference for offloading only
& \textcolor{green}{\cmark} Multi-level Hierarchical \& oversubscribed\\
& & & & \cellcolor{cyan!5} \\
\textbf{Memory\newline Modeling\newline} 
& \textcolor{red}{\xmark} Post-hoc check
& \textcolor{red}{\xmark} Post-hoc check (defaults to over-sharding to fit memory)
& \textcolor{green}{\cmark} Integrated memory optimizations
& \textcolor{green}{\cmark} Native; prunes invalid states and prevents over-sharding \\
& & & & \cellcolor{cyan!5} \\
\textbf{Scalability\newline} 
& \textcolor{orange}{\protect\ding{110}} Poor with large search spaces
& \textcolor{orange}{\protect\ding{110}} Poor beyond $\approx$64 GPUs
& \textcolor{orange}{\protect\ding{110}}  Poor with large search space
& \textcolor{green}{\cmark} Scales to 1000+ GPUs \\
& & & & \cellcolor{cyan!5} \\
\textbf{Expert \& ZeRO Support\newline} 
& \textcolor{red}{\xmark} Not supported 
& \textcolor{green}{\cmark} Integrated in Intra-Operator 
& \textcolor{orange}{\protect\ding{110}} ZeRO support only
& \textcolor{green}{\cmark} Native, adaptive support \\
& & & & \cellcolor{cyan!5} \\
\textbf{Optimality\newline Guarantees} 
& \textcolor{red}{\xmark} None (random search) 
& \textcolor{orange}{\protect\ding{110}} Heuristic, limited by network \& partitioning strategy
& \textcolor{orange}{\protect\ding{110}} Yes, limited by network \& partitioning strategy
& \textcolor{green}{\cmark} Structured optimality via DP \\
\bottomrule
\end{tabular}
\label{tab:comparison}
\vspace{-2.5ex}
\end{table*}

%% file: body/background.tex
\section{Background and Motivation}

\niparagraph{Distributed Deep Learning.}
Modern frameworks such as Megatron-LM~\cite{megatronlm}, NeMo~\cite{nemo}, and DeepSpeed~\cite{deepspeed} enable scalable training via multiple parallelization strategies.
Intra-layer tensor parallelism (TP) splits individual layers across multiple accelerators, with sequence parallelism (SP) further partitioning activations along the sequence dimension to reduce memory pressure. Expert parallelism (EP) distributes Mixture-of-Experts (MoE) modules across devices, introducing unique all-to-all communication patterns. Context parallelism (CP)~\cite{context_parallel} segments the input sequence across multiple GPUs, specifically to handle the quadratic memory scaling of long-context attention mechanisms. Pipeline parallelism assigns layers to different devices, enabling micro-batch processing to reduce per-device memory usage~\cite{gpipe, pipedream}, and data parallelism replicates the model across devices. Memory optimizations like ZeRO~\cite{zero} partition optimizer states, gradients, and parameters to reduce memory usage. Each approach introduces distinct communication patterns and resource trade-offs, requiring careful orchestration for efficient training.


\niparagraph{Datacenter Network Topologies.}
Real-world datacenters are rarely uniform, typically employing hierarchical topologies such as Fat-Tree, Spine-Leaf, or Clos networks to balance scalability, fault tolerance, and cost~\cite{msftmaia, superpod}. Practical deployments often feature oversubscription or non-uniform interconnects due to physical layout and cost constraints, resulting in widely varying latency and bandwidth across intra-node, intra-rack, and inter-rack links. For instance, same-rack communication may use full-bandwidth NVSwitch connectivity, whereas cross-rack transfers traverse spine switches with shared or oversubscribed bandwidth. These hierarchical non-uniformities make distributed training highly topology-dependent. Appendix~\ref{app:topology} illustrates common network configurations.

\niparagraph{Prior Device Placement Techniques.}
Many works have explored automated model partitioning using reinforcement learning~\cite{deviceplacement-rl}, Markov Chain Monte Carlo (MCMC)~\cite{topoopt}, and dynamic programming (DP)~\cite{pip, piper, phaze, alpa}. Early systems such as PipeDream~\cite{pipedream} focused primarily on pipeline parallelism, while later frameworks, including \cite{piper, alpa, mist, aesco}, combined data, operator, and inter-layer parallelism. However, most assume flat or simplified networks and do not fully support emerging strategies such as expert parallelism or memory optimizations like ZeRO.

In practice, datacenter networks are \textbf{hierarchical and often oversubscribed}, with latency and bandwidth varying across intra-node, intra-rack, and inter-rack links. These differences directly affect performance-critical collectives such as \texttt{AllReduce}, \texttt{AllGather}, and \texttt{ReduceScatter}, which frequently dominate distributed training overhead~\cite{astrasim, microChar}.

Several topology-aware approaches attempt to address these realities but remain limited in scalability or optimality. TopoOpt explores placements stochastically, lacks optimality guarantees, is sensitive to initialization, and scales poorly as the number of parallelization dimensions grows. Alpa uses dynamic programming, but to keep the search tractable, it assumes simplified 2D mesh networks that cannot capture oversubscription or multi-level hierarchy. It prioritizes minimizing communication without jointly modeling compute latency and checks memory feasibility only after generating placement plans (i.e., post hoc).
The lack of integrated memory modeling further increases search time and reduces the ability to find feasible placements on smaller clusters.
To meet memory budgets, Alpa aggressively shards parameters and activations across GPUs. On larger clusters, it optimizes each pipeline stage independently and constructs a single pipeline; additional devices are used to further shard layers rather than scale via pipeline replication. This can lead to hardware underutilization and excessive communication, limiting effective scaling beyond roughly 64 GPUs.

To address these limitations, more recent works shift their focus toward memory feasibility and scheduling, often at the expense of explicit network modeling. Aceso~\cite{aesco} uses greedy, bottleneck-chasing heuristics to improve memory balance and hardware utilization. However, its local search can stall in suboptimal configurations and does not provide global optimality guarantees. Mist~\cite{mist} formulates placement and scheduling as an MILP problem, emphasizing temporal overlap between communication and computation. While it handles memory constraints more robustly than Alpa, it treats network topology as a secondary consideration and relies on hierarchical brute-force search for parallelization, which limits scalability on large hierarchical clusters.

This contrast between prior works is showcased in Table~\ref{tab:comparison}.

\begin{figure}[t]
    \centering
    \includegraphics[width=1\columnwidth]{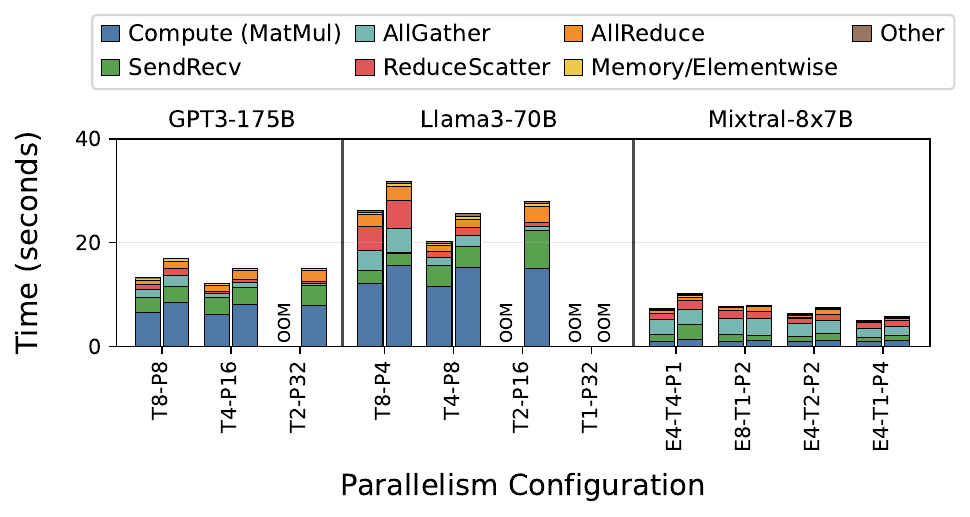}
    \vspace{-3ex}
    \caption{Impact of communication latency on training time across different parallelism strategies on an oversubscribed 64-GPU cluster; left bar (without) and right (with) activation recomputation.}
    \label{fig:motivation}
    \vspace{-3ex}
\end{figure}

\niparagraph{The Need for Topology-Aware Placement.}
As training scales to hundreds of accelerators, small differences in placement can cause large performance variations. Figure~\ref{fig:motivation} illustrates this for GPT3-175B, Llama3-70B, and Mixtral-8$\times$7B models on a 2:2 oversubscribed 64-GPU cluster, where communication accounts for a significant portion of total training time. The most efficient parallelization strategy depends on both the model and the underlying topology—what works on a uniform mesh may perform poorly on a hierarchical or bandwidth-asymmetric cluster. Efficient training, therefore, requires a placement framework that explicitly models network hierarchy, asymmetry, and oversubscription.

\niparagraph{Motivation for \nest.}
These limitations highlight the need for a placement framework that \textbf{(i) Strategy-aware}, co-optimizing data, tensor, pipeline, expert, sequence,  context, and memory parallelism; \textbf{(ii) Network-aware}, explicitly modeling hierarchical, oversubscribed, and asymmetric interconnects; and \textbf{(iii) Scalable and principled}, using structured optimization like dynamic programming rather than heuristics or sampling.

We introduce \textbf{\nest}, a network- and memory-aware dynamic programming framework that addresses these gaps. Unlike prior works that rely on stochastic search or simplified topology models, \nest explicitly models hierarchical, oversubscribed, and heterogeneous networks while integrating memory-modeling into its search methodology. 
This enables \nest to systematically explore feasible placements that balance communication, compute, and memory efficiency, scaling effectively across heterogeneous and oversubscribed datacenter architectures.

%% file: body/design.tex
\begin{figure*}[ht]
\centering
\includegraphics[width=\textwidth]{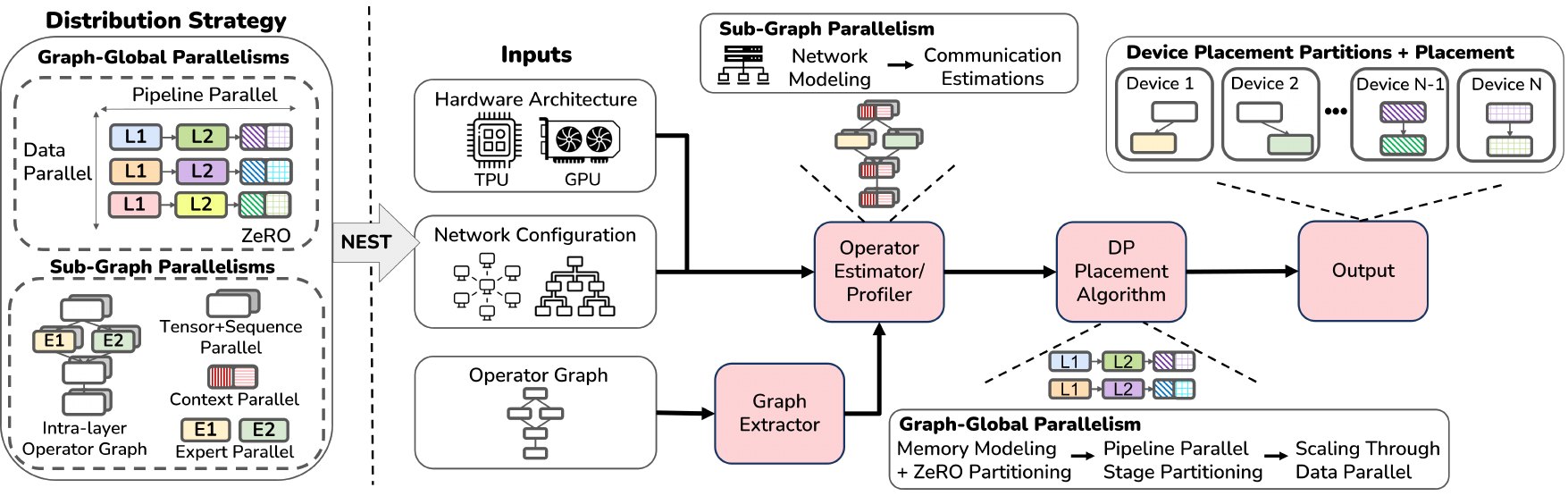}
\vspace{-2ex}
\caption{\nest~search space and workflow. \graphglobal strategies partition entire layers, whereas \subgraph strategies partition computations within individual layers. Different colors denote device assignments, and dashed boxes show ZeRO and context parallelism.}
\vspace{-2ex}
\label{fig:overview}
\end{figure*}


\section{\nest Overview}

\nest~jointly models compute, memory, and communication during the search process to optimize network-aware device placement and distributed parallelization strategy for large-scale training under realistic data center conditions. 
Figure~\ref{fig:overview} illustrates the \nest framework and the representative set of parallelization strategies supported by it.

\subsection{Categorization of Parallelization Strategies}

A key insight behind \nest~is that distributed training parallelization techniques can be organized along \emph{orthogonal dimensions} based on their interaction with the model graph and hardware.
This principled categorization makes the dynamic programming search tractable by separating local, intra-operator transformations from global, inter-operator scheduling decisions.
It enables \nest~to compose strategies such as tensor, pipeline, data, expert, sequence, and context parallelism into hybrid configurations without combinatorial growth of the search space.
\emph{Factoring optimization along these dimensions yields a systematic, extensible framework: new parallelization methods can be integrated seamlessly, and hybrid placements explored efficiently within a unified dynamic programming formulation.}
Building on this principle, \nest~classifies strategies into two categories: \subgraph and \graphglobal, each corresponding to a distinct abstraction level in the computation graph.

\niparagraph{\subgraph strategies.}
%
These strategies operate within individual operators or small subgraphs, such as tensor and expert parallelism. 
\nest~classifies a strategy as \subgraph if it modifies an operator’s \emph{internal execution}, e.g., by partitioning parameters or adjusting tensor dimensions per accelerator, while preserving the model’s overall control and dataflow (i.e., layer sequence and dependencies).
Its goal is to alleviate compute and memory bottlenecks within a layer by splitting its workload (e.g., weight matrices or expert activations) across devices. This fine-grained partitioning introduces tightly coupled collective operations, such as \texttt{AllReduce} for tensor parallelism, \texttt{AllToAll} for expert parallelism, and \texttt{AllGather} and \texttt{ReduceScatter} for sequence and context parallelism to synchronize results.
Because these transformations are local and self-contained, \nest~profiles their operator-level compute, memory, and communication costs offline. These pre-characterized costs are then composed analytically during higher-level placement, without expanding the dynamic programming (DP) search space. 
Any operator-level technique replacing a layer with an equivalent distributed implementation can be integrated as a \subgraph strategy by providing its transformed computation graph and runtime annotations.

\niparagraph{\graphglobal strategies.}
\graphglobal~strategies act over the entire model computation graph, such as pipeline parallelism (partitioning layers into stages), data parallelism (replicating the full model and synchronizing gradients), and ZeRO-style sharding (distributing optimizer states).
They decide \emph{how layers, parameters, and data batches are distributed and scheduled} across multiple devices or nodes, thereby reshaping the global execution plan.  
%
%
These strategies affect the global communication structure, memory balance, and inter-stage dependencies, and their impact cannot be localized to individual operators.
\nest~therefore incorporates \graphglobal~strategies directly within its novel DP-based search, where it optimizes stage boundaries, replication factors, memory footprint, and communication overhead jointly under system-level constraints.

\niparagraph{Discussion.}
%
By treating \subgraph and \graphglobal strategies as orthogonal dimensions, with local transformations and global scheduling, \nest avoids the combinatorial explosion of enumerating all configurations.
Instead, it composes pre-profiled local strategies within a globally optimized placement, enabling systematic integration of new parallelism forms (e.g., expert or activation sharding) under a unified compute–memory–communication cost model.

\subsection{\nest Workflow}

NEST constructs a unified optimization workflow across three stages: \emph{graph extraction}, \emph{runtime estimation}, and \emph{placement search}. Importantly, \nest operates strictly as a \textbf{planning system}: it preserves the mathematical equivalence of the original training formulation and does not modify the underlying model code or kernels. 

It takes as input (1) a hardware specification (e.g., GPUs, TPUs, or domain-specific accelerators), (2) the model’s operator graph, and (3) a detailed network configuration specifying topology, bandwidth, latency, and communication protocols.  

\niparagraph{Graph Extraction.}
\nest extracts the operator graph from the training script using symbolic tracing. It then applies user-specified or predefined logical \subgraph transformations (e.g., tensor or expert splits) and inserts the required collective communication operators at the appropriate locations. This produces an operator graph that represents the distributed execution of the original model. Each resulting variant is a self-contained graph used for runtime modeling.

\niparagraph{Runtime Estimation.}
For each extracted graph, \nest performs an offline analysis to annotate compute and communication operators with platform-specific runtimes. Compute costs are derived from hardware estimators or profilers such as PyTorch~\cite{pytorch}, while communication costs are estimated using network simulators like AstraSim~\cite{astrasim}. This provides a unified view of computation, communication, and memory behavior across the hierarchy.

\niparagraph{Solver.} 
Given operator graphs annotated with compute and communication costs, per-layer memory usage, and cluster network characteristics, the DP solver explores \graphglobal configurations, including pipeline stage divisions, replication degrees, and data partitions, while incorporating the cost models of \subgraph strategies. This search evaluates trade-offs among latency, memory, and bandwidth to identify placements that maximize end-to-end throughput under memory and communication constraints. The final output is a parallelism configuration and placement plan.

The categorization of parallelization strategies allows \nest to produce scalable and generalizable device placements. By leveraging established parallelization strategies, the resulting plan runs on standard distributed frameworks such as Megatron-LM~\cite{megatronlm} and NeMo~\cite{nemo} while preserving the mathematical equivalence of the original training formulation.

\subsection{\nest Memory Modeling}

Following prior work~\cite{phaze, mist, nnscaler}, \nest uses symbolic analysis via \texttt{Torch.fx}~\cite{torchfx} to estimate workload memory requirements. During graph extraction, \nest traces the dependency graph and annotates operator metadata, including parameters, input sizes, and activation tensors. This is then used by the Solver for memory modeling.

The peak memory footprint of a contiguous subgraph (stage) $S$ is modeled during the pipeline’s “steady state” and includes five components: model weights, accumulated gradients, optimizer states, current intermediate activations, and stashed data (activations held for in-flight microbatches). The amount of stashed data depends on the schedule: in 1F1B, a stage at index $s$ from the pipeline end holds $(s-1)$ microbatches; in GPipe, this scales by $B/d$.

\nest evaluates two Activation recomputation (AR) strategies: 
(1) \textbf{Without Activation Recomputation}, where all intermediate activations are stashed; and 
(2) \textbf{With Activation Recomputation}, where only stage-boundary input activations are stashed, with intermediate tensors re-materialized during the backward pass. 

\nest calculates peak memory using the equation:
{
\begin{multline}
\footnotesize
\text{Mem}(S, s) = \\
\sum_{L_i \in S} \Big( 2 \cdot \text{weights} + \text{opt\_states} + \text{activations} \Big) + \\
(s-1) \cdot \text{stashed\_data}
\label{eq:nest_memory}
\end{multline}
}

By modeling memory as a linear function of stage position $s$, the solver avoids redundant computations across different pipeline positions, which significantly speeds up the search. \nest’s memory estimates were validated against compiled kernels and are on average within 7\% of actual usage across evaluated models. A summary of the model estimations is in Appendix~\ref{app:memory}.

\section{\nest's Dynamic Program}
\label{sec:dp}
\nest~introduces a novel network-, compute-, and memory-aware dynamic programming (DP) formulation that directly embeds network topology characteristics, bandwidth asymmetry, and per-stage compute latency and memory constraints into the optimization process. Even though prior works have employed DP for device placement, none have integrated network, compute, and memory awareness into a unified optimization framework~\cite{alpa, phaze, piper}. By doing so, \nest can reason about co-location, overlap, and pruning jointly, ensuring feasible and high-efficiency placements across heterogeneous clusters. This section presents the design of this DP solver, evolving from a baseline formulation to a fully integrated, topology- and memory-conscious optimizer.

\niparagraph{DP Inputs and Search Space.}
\nest\!’s DP takes as input profiled operator graphs capturing multiple candidate \subgraph parallelism configurations. Each configuration records operator compute latency, collective communication costs (e.g., \texttt{AllReduce}, \texttt{AllToAll}), and per-layer memory usage. The DP also accounts for memory-saving techniques such as ZeRO sharding (parameters, gradients, optimizer), pipeline parallelism for layer partitioning, and data parallelism for scalable replication.

These profiled graphs and system characterizations define the DP \textbf{search space}. 
Each state represents a partial assignment of layers, devices, and parallelization strategies, while each DP transition is a feasible extension respecting memory and network constraints. 
By reasoning over this structured search space, \nest systematically explores thousands of hybrid placements without explicit enumeration, balancing compute distribution, communication locality, and memory feasibility within a unified optimization framework.

\niparagraph{Local vs. Global Parallelism in the DP.}
\nest separates \subgraph (local) and \graphglobal strategies within its DP formulation.
Local strategies, such as tensor and expert parallelism, modify how a layer’s internal computation and parameters are distributed across accelerators. Their effects are captured during profiling and influence the per-stage latency term $\text{load}(\cdot)$.
In contrast, \graphglobal strategies, such as data, pipeline, and ZeRO parallelism, reshape the boundaries between layers or stages and are explicitly explored by the DP during partitioning and placement.
This separation allows \nest to maintain a tractable state space while still enabling hybrid configurations (e.g., tensor + pipeline + data parallelism) to be composed systematically.
\subgraph strategies incorporate network awareness by modeling collective communication at multiple locality levels, such as intra-node, inter-node, and inter-rack, allowing the DP to reason about how local communication costs vary with physical placement and topology.

\niparagraph{Baseline DP: Network-Agnostic.}
We start with a simplified baseline that ignores network and memory constraints.  
Let $\texttt{dp[D][k][s]}$ denote the minimum latency of executing a downset $D$ (a suffix of the layer graph) across $k$ devices and $s$ pipeline stages.  
The recurrence enumerates all subgraphs $D' \subseteq D$ and possible device allocations $a$, considering \mbox{\subgraph} strategies ($sg$) such as tensor, expert, sequence, and context parallelism:  
{
\vspace{-1ex}
\begin{multline}
\small
dp^{sg}[D][k][s] =  \min_{D' \subseteq D} \min_a \max \\ \Big( dp^{sg}[D'][k-a][s-1], \;
\text{load}^{sg}(D \setminus D', a, s) \Big)
\label{eq:origdp} 
\end{multline}
}
Here, $D \setminus D’$ denotes the new stage assigned to $a$ devices, while $\text{load}^{sg}(S,a,s)$ estimates the latency of that stage given the chosen \subgraph parallelism configuration. The recurrence then proceeds recursively, optimizing the remaining layers over the remaining $k-a$ devices and $s-1$ stages.
This formulation efficiently prunes suboptimal decompositions but assumes a fully connected, uniform-cost network and ignores memory-induced constraints.

\niparagraph{Challenge and Level-Wise Network Abstraction.}
When network heterogeneity is introduced, a core challenge emerges: each pipeline stage’s latency depends not only on its own compute and backward communication, but also on the \emph{unknown placement of its predecessors} that produce forward activations.
Since the DP proceeds backward (from last to first stage), the producer’s location, and thus its communication cost, is unknown at decision time.

\begin{figure}[t]
    \centering
    \includegraphics[width=1\linewidth]{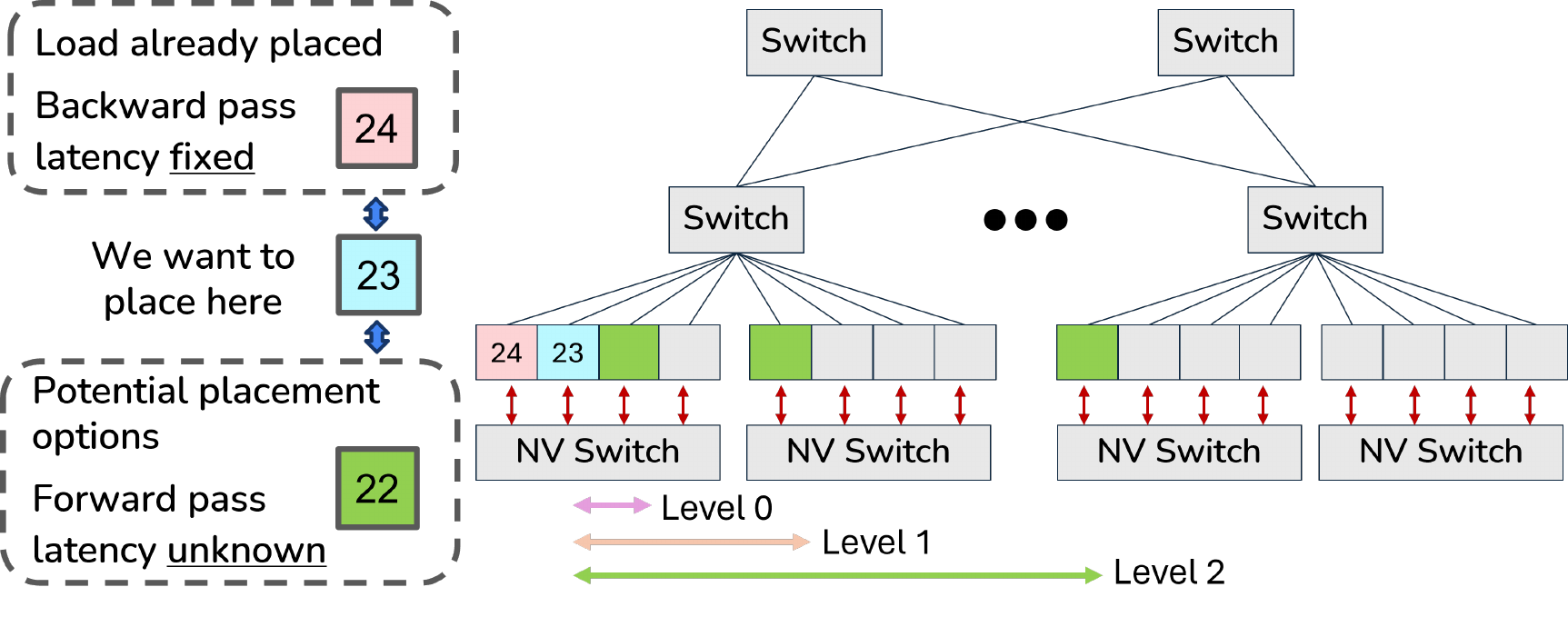}
    \vspace{-2.6ex}
    \caption{The forward-pass dependency challenge and \nest\!'s level-wise abstraction. The cost from unassigned Layer 22 to Layer 23 is unknown but abstracted as a discrete communication level (e.g., Level 0, 1, or 2).}
    \label{fig:comm-problem}
    \vspace{-3ex}
\end{figure}

Figure~\ref{fig:comm-problem} illustrates this challenge. When placing layer 23, the backward link to layer 24 (already placed) has a known latency based on their relative placement, but the forward link from layer 22 depends on where that stage will be located (e.g., same node, same rack, or remote). This asymmetry breaks the optimal substructure assumption.

\niparagraph{Level-Wise Network Abstraction.}

To restore tractability, \nest introduces a \textbf{level-wise abstraction}  that groups devices by communication locality. Instead of modeling every device pair, \nest treats each network level $l$ as a discrete DP state. This lets \nest generalize to real-world topologies by building a level-wise cost matrix based on physical interconnects. Examples include:

\begin{itemize}[leftmargin=*, itemsep=1pt, parsep=0pt, topsep=1pt, partopsep=1pt]
    \item \textbf{Hierarchical Fabrics (NVIDIA HGX/Spine-Leaf/Fat Tree):} Levels correspond to switch tiers and link types. For example, $l_0$ represents intra-node NVLink (e.g., 900 GB/s for H100), while $l_1$ and $l_2$ capture the bandwidth differences of inter-node Ethernet/InfiniBand and oversubscribed inter-rack Spine-Leaf links. This prevents the DP from placing high-bandwidth pipeline stages across oversubscribed boundaries.
    
    \item \textbf{Non-Uniform Meshes (TPUv4 Torus):} In torus-based systems, levels represent hop distance or coordinate-affinity classes (e.g., $l_0$: same tile, $l_1$: 1-hop, $l_2$: remote). Mapping these to profiled latencies lets \nest account for non-uniform communication costs when scaling to thousands of accelerators.
\end{itemize}

As shown in Figure~\ref{fig:comm-problem}, the unknown forward cost is captured by just a few network levels (typically 3–5). This allows the DP to reason over levels instead of all device pairs, reducing combinatorial complexity while preserving hierarchical topology fidelity. Appendix~\ref{app:network_mesh} shows how this abstraction generalizes to non-uniform mesh and torus architectures.

\niparagraph{Key Observation.}
This abstraction is topology-agnostic. By decoupling locality from a fixed hierarchy, the level-wise abstraction allows \nest's DP to generalize to diverse and future interconnects, preserving scalability while adapting to new topologies and communication models.

\niparagraph{Proposed DP: Network-Aware.}
With this abstraction, the DP recurrence becomes:
\begin{multline}
\small
dp^{sg}[l][D][k][s] = 
\min_{D' \subseteq D} \min_a 
\max \\ \Big( 
    dp^{sg}[l'][D'][k-a][s-1], \;
    \text{load}^{sg}_l(D \setminus D', a, s) 
\Big)
\label{eq:networkdp}
\end{multline}

Here, $dp^{sg}[l][D][k][s]$ denotes the minimum possible latency (i.e., the cost of the bottleneck stage) to execute the suffix of layers $D$ using $k$ devices partitioned into $s$ pipeline stages under \subgraph parallel configurations $sg$. The key new term is $l$ (level). Since the DP proceeds backward, from the last layer toward the first, the placement of the next stage (e.g., layer 22 in Figure~\ref{fig:comm-problem}) is unknown when placing the current stage (e.g., layer 23). $l$ represents the assumed communication distance of the yet-unplaced producer stage relative to the earliest layer in $D$, acting as a ``deferred forward cost'' that preserves optimal substructure.

\niparagraph{Unified Cost Model and Recurrence.}  
Equation~\ref{eq:networkdp} jointly finds the optimal pipeline cut-point ($D'$) and device allocation ($a$) for the new stage ($D \setminus D'$), exploring transitions across communication levels while pruning memory-infeasible states. The latency estimate $\text{load}^{sg}_l(\cdot)$ is central to co-optimization, capturing:  
\begin{itemize}[leftmargin=*, itemsep=1pt, parsep=0pt, topsep=1pt, partopsep=1pt]
\item \textbf{Compute latency:} derived from profiled per-operator runtimes, scaled by \subgraph parallelism ($sg$) degrees.
\item \textbf{Network latency:} determined by the level-wise communication cost matrix, accounting for collectives and both forward (from level $l$) and backward (to $D'$) traffic.
\item \textbf{Memory-Optimization Co-design:} AR and ZeRO are natively integrated into the state transition. If a state exceeds device memory, the solver incrementally increases ZeRO levels (1, 2, or 3) until feasibility is reached, adding the resulting collective overhead to the latency. Similarly, recomputation is modeled as a binary optimization choice that reduces the "stashed data" memory term at the expense of higher compute latency.
\end{itemize}

The recurrence uses $\max(\cdot, \cdot)$ to balance the new stage's cost ($\text{load}^{sg}_l$) with the remaining stages ($dp^{sg}[l'][D'][k-a][s-1]$), where $l'$ encodes the communication level between consecutive stages. This unified model ensures \nest considers only valid configurations, automatically trading off memory reduction (sharding) against communication overhead (co-location). Overall, this formulation allows \nest to jointly optimize placement, communication, and memory, favoring co-location of frequently communicating stages while penalizing cross-level traffic.

\niparagraph{Example.}
Consider the example in Figure~\ref{fig:comm-problem}. When placing layer 23, the DP enumerates its possible locations at each communication level relative to layer 24. For each option, it pre-computes the deferred forward costs corresponding to potential placements of layer 22. These partial sub-solutions are stored in \texttt{dp[l][{23,24}][2][2]}. Later, when placing layer 22, \nest simply uses these precomputed costs to select the globally optimal configuration—achieving optimality without exhaustive enumeration.

\niparagraph{Summary.} 
\nest's dynamic programming is novel in three ways: (1) a hierarchical level-wise abstraction making network-aware placement tractable, (2) a unified compute–network–memory model embedding feasibility in the recurrence, and (3) orthogonal handling of local and global parallelism within a structured optimization framework.

%% file: body/evaluation.tex
\section{Evaluation}
\label{sec:eval}

\subsection{Methodology and Setup}
\label{sec:setup}

\niparagraph{Models:} Table~\ref{tab:workload} lists the large language models used to evaluate \nest, including Llama2-7B~\cite{llama2}, Llama3-70B~\cite{llama3} (Hugging Face~\cite{huggingface}), and BertLarge~\cite{bert}, GPT3-175B, and Mixtral-8x7B~\cite{gpt3,mixtral} (tensor and expert parallelism; Megatron-LM~\cite{megatronlm, shoeybi2020megatronlm}). Operator graphs are extracted with Torch.fx~\cite{torchfx}. A global batch size of 4096 and a microbatch size of 1 are used for all experiments, unless stated otherwise.

\niparagraph{Operator-Level Estimates, Runtime Profiles, and \nest Execution:} For TPUv4-like accelerators, operator latencies are estimated using hardware-validated libraries Sunstone~\cite{sunstone} (tensor cores) and Tandem~\cite{tandem:asplos:2024} (vector cores), with ILP-based layer scheduling from prior work~\cite{phaze}. For GPUs, compute latencies are profiled at operator and layer levels using the PyTorch Profiler~\cite{pytorch}, and communication collectives are modeled and validated with AstraSim~\cite{astrasim} (Appendix~\ref{app:val}). \nest runs on an H100 GPU to extract operator graphs and runtime profiles. Experiments use RHEL~9.6, Python~3.10, and a C++ solver compiled with \texttt{g++}~12.4.0.

\niparagraph{Baselines.}
We compare \nest against five baselines:
(1) Manual placements from prior work~\cite{efficient_Megatron, phaze}, scaling data parallelism across cluster sizes.
(2) Phaze~\cite{phaze}, a network-unaware DP framework built on Piper~\cite{piper}; we evaluate only its device placement component.
(3) MCMC-based placement~\cite{topoopt}, implemented to explore the same parallelization strategies as \nest.
(4) Alpa~\cite{alpa}, a state-of-the-art DP framework leveraging intra-operator parallelism.
(5) Mist~\cite{mist}, a state-of-the-art DP framework focusing on scheduling and memory optimizations.

For fairness, \nest and baselines use PipeDream-Flush schedule~\cite{pipedream-flush} and shared cost model.  All MCMC baselines were run 10 times, reporting the best-performing result to ensure robust comparison. A key challenge with Alpa at scale is that its native design requires full-cluster access for runtime profiling, which is infeasible for large-scale deployments. To address this, we create an offline variant, Alpa-E (Alpa Estimator), which retains Alpa’s core optimization while replacing its hardware-dependent profiler with our unified estimator. This isolates differences in search efficiency and placement quality under identical cost assumptions. In Section~\ref{sec:spine_alpa}, we evaluate Alpa’s original profiling-based implementation, Alpa-O, on real hardware where direct profiling is feasible.

Similarly, the Mist placement search algorithm requires execution on at least one physical device to function, and its current implementation is optimized specifically for NVIDIA architectures. Consequently, we evaluate Mist exclusively in Section~\ref{sec:spine_large} during our large-scale experiments on NVIDIA H100 GPUs. This ensures that the baseline is evaluated within its intended operational environment. In this evaluation, we compare against Mist’s placement strategy rather than its scheduling optimizations, as we consider those optimizations orthogonal to \nest’s goals. We expect even larger gains if \nest’s network-aware placement is combined with Mist’s scheduling algorithm.

\input{body/fatTree_table}

\begin{figure*}[h!]
\center
\vspace{-2ex}
\includegraphics[width=0.98\textwidth]{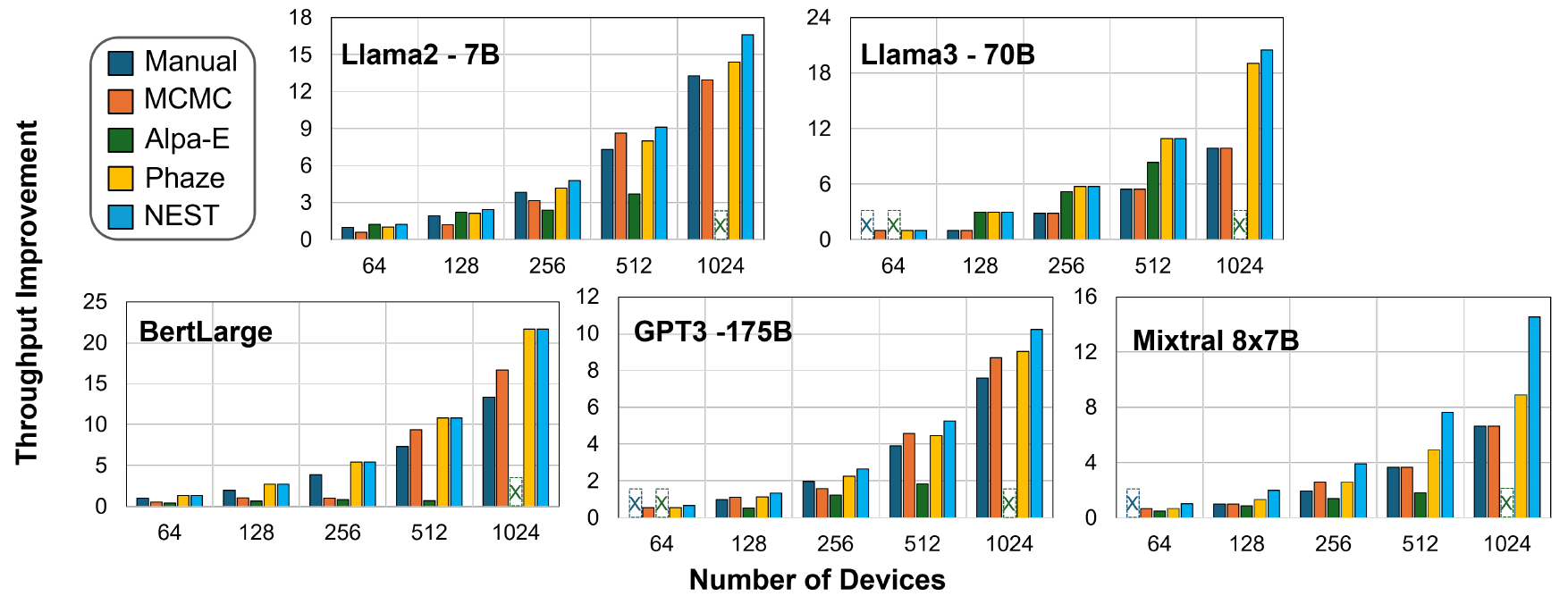}
\vspace{-2ex}
\caption{Throughput comparison between \nest~and baselines on a Fat-Tree network of TPUv4 accelerators. Throughput improvements are relative to the manual baseline’s smallest valid result. “X” indicates cases where the baseline failed to find a valid placement.}
\label{fig:throughput}
\vspace{-1ex}
\end{figure*}

\subsection{Accelerators with Fat-Tree Execution}
\label{sec:fat_large}
We evaluate \nest on fat-tree topologies with 64–1,024 TPUv4-like accelerators. Each node has eight accelerators connected via an HGX-style link (900 GB/s), with four nodes per first-level switch (100 GB/s) and second-level aggregation at 400 GB/s (Figure \ref{fig:tree}). Table \ref{tab:workload} lists workloads, and Figure \ref{fig:throughput} reports throughput gains.
Alpa is limited to 512 devices due to profiling overhead, which grows rapidly as it enumerates all layer–mesh configurations (up to 48 hours, and in some cases doesn't converge in 3 days). In contrast, \nest completes optimization in 3 minutes to 1.5 hours, scaling from BertLarge to GPT-3 175B on 1,024 devices, by constructing valid subgraphs per device and using template-based parallelism. MCMC and Phaze show similar runtimes.
We further study the impact of ZeRO Optimization in Appendix~\ref{app:zero} under a similar network topology.

\subsubsection{Throughput Improvements}

On average, \nest’s distribution strategy achieves \textbf{1.59$\times$ higher throughput than manual placements}, \textbf{1.71$\times$ higher than MCMC-based search}, \textbf{2.43$\times$ higher than Alpa-E}, and \textbf{1.19$\times$ higher than Phaze}. Importantly, \nest scales nearly linearly with cluster size, while other baselines either fail to find valid placements or experience diminishing returns. These results demonstrate the effectiveness of \nest’s network-aware optimization and systematic search in delivering both high performance and robust scalability.

\niparagraph{Comparison with Random Search Methods (MCMC).}
MCMC-based methods rely on random exploration and offer no optimality guarantees, with performance degrading as cluster size and model scale grow. While small models like BertLarge may see competitive strategies, larger models such as Megatron-GPT3 and Mixtral-8x7B require far more iterations for meaningful gains. Even after an extensive search, MCMC underperforms both manual placements and \nest, and is highly sensitive to initialization. For example, on Llama2-7B, MCMC matches \nest only at 512 devices but fails at other scales; GPT3 and Llama3-70B show similar trends. These results motivate \nest: the need for structured, deterministic optimization that efficiently explores the parallelization space without exhaustive search.

\begin{figure*}[t]
    \centering
    \includegraphics[ width=1\linewidth]{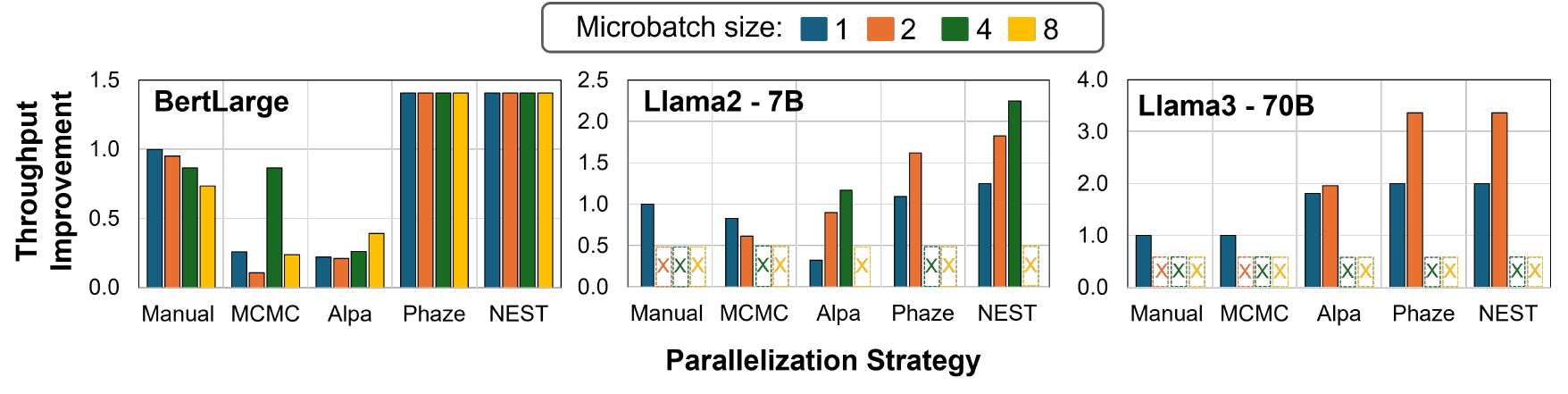}
    \vspace{-4ex}
    \caption{Throughput improvement relative to a manual baseline (microbatch size 1). “X” marks baseline failures due to memory constraints. High memory requirements limit Llama2-7B and Llama3-70B to microbatch sizes 4 and 2, respectively.}
    \label{fig:batch}
    \vspace{-2ex}
\end{figure*}

\niparagraph{Comparison with Phaze.}
Phaze’s dynamic programming (DP) assumes a flat, uniform network. While it balances computation, it overlooks communication costs, reducing throughput on heterogeneous interconnects—especially at scale.
In contrast, \nest incorporates network topology and bandwidth heterogeneity directly into its DP solver, selecting parallelism degrees and stage assignments that minimize communication while balancing computation across an expanded search space that includes SP and CP. For example, in Llama2-7B and Mixtral-8×7B, \nest\!’s network-aware partitioning improves pipeline balance and reduces stalls.
Even under a similar parallelism plan (e.g., GPT3-175B), \nest achieves higher throughput by assigning fewer layers to stages connected via slower links, equalizing total stage time (compute + communication). This topology-adaptive balancing mitigates pipeline bubbles and link bottlenecks, sustaining near-linear scaling as models and clusters grow.
When the model fits within a single pipeline stage ($p=1$) or each layer forms its own stage ($p = \#L$), Phaze and \nest achieve similar throughput (e.g., BertLarge).

\noindent\fbox{%
\parbox{\columnwidth}{%
\textbf{Insight: Topology-adaptive partitioning} enables \nest to equalize stage latency across heterogeneous links, maintaining balanced pipelines and sustaining linear scaling where topology-agnostic baselines stall.}}

\niparagraph{Comparison with Alpa.}
At small scales ($\leq$128 devices), Alpa-E achieves throughput comparable to \nest, with gains up to 3\%, stemming from its fine-grained operator and layer sharding. On larger clusters and communication-heavy models like GPT-3-175B and Mixtral-8×7B, Alpa’s performance drops due to three main limitations: (1) memory feasibility is checked only post placements, (2) pipeline stages are optimized independently, limiting data-parallel scaling, and (3) Alpa’s 2D mesh assumes uniform communication, ignoring hierarchical network effects, which leads to over-sharding and suboptimal placements at larger scales.

\textit{Memory Modeling.}
Alpa verifies memory feasibility post hoc, often forcing aggressive sharding of activations and parameters, increasing search time and  potentially failing on large models for small clusters (e.g., GPT3-175B or Llama3-70B on 64 GPUs). In contrast, \nest integrates memory and network constraints into its dynamic programming, evaluating each subgraph placement against memory budgets. Combined with ZeRO-based subgraph decomposition, this enables efficient scaling of large models on fewer devices.


\textit{Effects of Over-sharding.}
Alpa optimizes stages independently, scaling larger clusters by splitting layers across devices rather than replicating pipelines, increasing communication and reducing utilization, especially for smaller models like BertLarge or clusters beyond 128 devices. \nest avoids this by first optimizing intra-stage efficiency and then replicating pipelines to fully utilize hardware while balancing throughput and communication. This enables partial cluster utilization when beneficial, whereas Alpa enforces full device usage even when it lowers per-device efficiency.


\noindent\fbox{%
\parbox{\columnwidth}{%
\textbf{Insight: Explicit memory-aware optimization.} Accurate memory modeling enables \nest to detect bottlenecks early and apply targeted optimizations, making training feasible where memory-limited baselines fail.}}




\subsubsection{Intra-operator vs Template-based Parallelism}

Alpa uses fine-grained intra-operator sharding to maximize parallelism but incurs high communication overhead, limiting performance on larger or heterogeneous clusters (beyond 128 devices). \nest and Phaze use template-based parallelism, such as tensor and expert parallelism, which exploit the repetitive structure of Transformer models with coarser granularity and lower communication. Phaze, however, assumes a flat interconnect and ignores network topology and bandwidth asymmetry, limiting performance at scale (e.g., 1,024 devices).
\nest overcomes these limitations by integrating network-aware placement into its template-based design, co-optimizing stage partitioning and inter-device communication based on real topology and bandwidth. This ensures balanced stage execution and avoids communication stalls, allowing \nest to match Alpa and Phaze on small clusters while delivering increasing throughput gains as cluster size grows.

\subsubsection{Joint Exploration with Microbatch Size}

Microbatch size and parallelization strategy strongly affect training throughput. \nest incorporates runtime factors like microbatch size and activation recomputation into graph extraction and cost estimation, enabling joint optimization with \subgraph parallelism.
Figure~\ref{fig:batch} shows throughput for \nest and baselines on BertLarge, Llama2-7B, and Llama3-70B across a 256-device cluster; 512-device results in Appendix~\ref{app:batch} show similar trends, except Alpa scales poorly due to oversharding.

Optimal microbatch size varies by model and strategy. For instance, while Llama models benefit from larger microbatches, BertLarge shows little improvement under \nest and even degrades under manual placement. Changes in microbatch size shift compute intensity and memory footprint, often altering the optimal parallelism configuration (e.g., Llama2-7B shifts from $\{P\text{=}8, D\text{=}64, T\text{=}1\}$ to $\{16, 32, 1\}$ as batch size increases from 1 to 2).
Alpa-E also benefits from larger microbatches, improving throughput by up to 1.8$\times$ for small models like BertLarge, but requires over 80 hours to sweep four batch sizes. In contrast, \nest completes the same joint exploration in 50 minutes, over 90$\times$ faster, while achieving consistently higher throughput, making comprehensive joint optimization practical at scale.

\begin{figure}
    \centering
    \includegraphics[ width=1\linewidth]{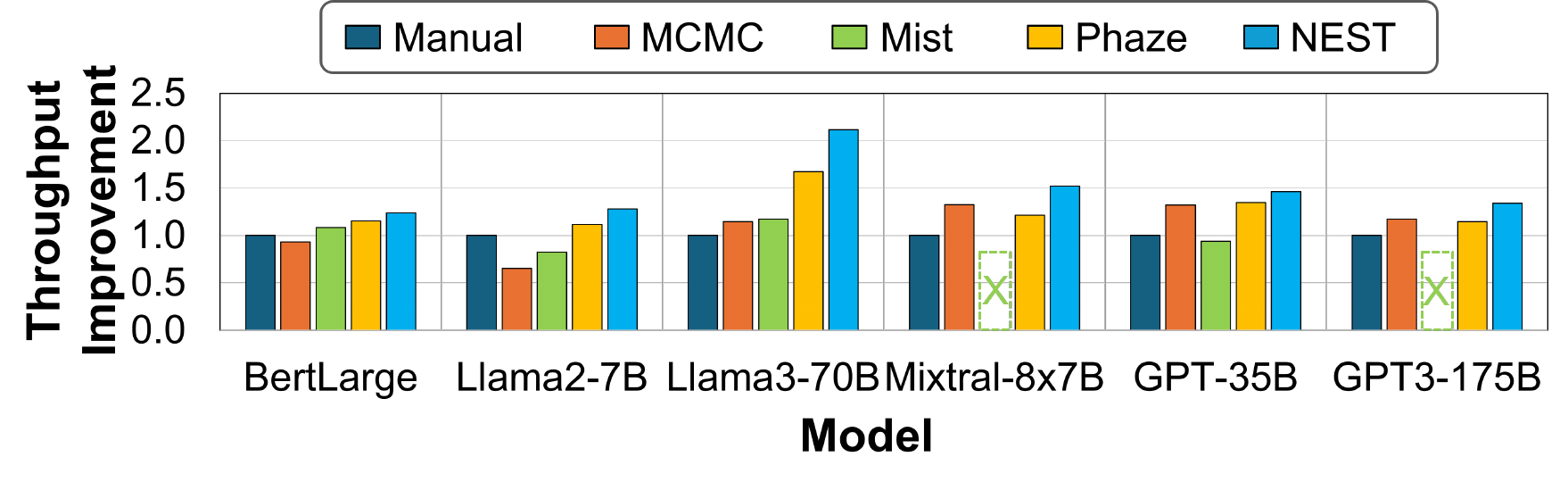}
    \vspace{-5ex}
    \caption{Throughput comparison between \nest~and baselines on a Spine-Leaf network of 1024 H100 GPUs. Throughput improvements are relative to the manual baseline. Mist does not support GPT3-175B and Mixtral-8x7B, as indicated with an "X".}
    \label{fig:h100}
    \vspace{-2ex}
\end{figure}

\subsection{H100 GPU Based Spine-Leaf Execution}
\label{sec:spine_large}

To demonstrate \nest’s generality across hierarchical topologies, we evaluate it on a 1,024-GPU H100 spine–leaf cluster based on the real topology also used in Figure~\ref{fig:motivation}. Each node has eight H100-80GB GPUs (NVLink 900~GB/s); first-level switches connect four nodes at 12.5~GB/s to two spine switches, forming a 2:2 oversubscribed topology. Operator runtimes are profiled on H100 GPUs, and collective communication costs are modeled with AstraSim.

We exclude Alpa-O from this evaluation because it requires full-cluster access for profiling, which is impractical in our setting. Although Alpa-E provides a cost estimation mode, using it would be inconsistent with our profiled H100-based setup and would not enable a fair comparison. Instead, we present direct hardware-based comparisons with Alpa-O on smaller clusters in Section~\ref{sec:spine_alpa}. For this evaluation, we compare against Mist~\cite{mist}, a state-of-the-art DP framework that focuses on scheduling and memory optimizations and requires access to only a single H100 for strategy tuning.


We compare against the same set of models listed in Table~\ref{tab:workload}, along with a scaled-down GPT-3 model (GPT3-35B; details in Appendix~\ref{app:gptsmall}). We use this smaller variant because Mist does not support GPT models with a hidden dimension larger than 8192, as used in the standard GPT3-175B configuration. In addition, Mist does not support MoE models such as Mixtral. Figure~\ref{fig:h100} shows the resulting throughput improvements.
On average, \nest’s distribution strategy achieves \textbf{1.47$\times$ higher throughput than manual placements}, \textbf{1.40$\times$ higher than MCMC-based search}, \textbf{1.49$\times$ higher than Mist}, and \textbf{1.16$\times$ higher than Phaze}. These gains stem from its ability to align computation stages with available bandwidth, delivering robust and scalable performance across diverse hardware and network configurations.



Both hardware and network variations strongly influence training performance, determining whether computation or communication becomes the bottleneck. For example, in the Mixtral model, communication can account for up to 10\% of total execution time on a constrained network, compared to only 1\% on a fat-tree topology. \nest distinguishes itself by explicitly modeling topology and communication costs during workload partitioning, enabling it to balance computation and communication effectively.
In contrast, Phaze’s network-agnostic strategy can cause severe latency imbalances when bandwidth is limited, and MCMC-based random search often fails in oversubscribed settings. For example, for the BertLarge model, Phaze selects $\{P\text{=}13, D\text{=}78, T\text{=}1\}$, leading to frequent cross-node communication and up to 2$\times$ imbalance in per-stage latency.
By considering the network, \nest instead selects $\{P\text{=}8, D\text{=}128, T\text{=}1\}$, keeping pipeline communication within a single node. This reduces inter-stage latency variation to under 2\% and consistently achieves higher throughput. This enables \nest to deliver stable performance gains and strong generalization across different hardware and network scales.


\niparagraph{Comparison with Mist.} \nest achieves higher throughput across all evaluated models thanks to its network-aware stage partitioning and placement strategy. For most models, \nest favors shallower pipelines to reduce pipeline bubbles and minimize cross-node communication, while using higher data-parallel degrees to scale throughput.
Mist supports uneven layer partitioning across pipeline stages, mainly to optimize memory and maximize compute-communication overlap. However, because it lacks network awareness, Mist cannot account for bandwidth differences across links and often fails to balance the communication-to-compute ratio across the cluster. In contrast, \nest’s topology-adaptive approach keeps stage latencies balanced, even when crossing oversubscribed rack links.

\nest also shows much higher search efficiency than Mist. Across all evaluated models, the \nest solver finds optimal placements on average 30\% faster under identical model and hardware configurations. A breakdown of exploration time for each model is provided in Appendix~\ref{app:runtime}.

\subsection{V100-GPU-Based Spine–Leaf Execution}
\label{sec:spine_alpa}

We validate \nest on real hardware using 8- and 16-GPU clusters (2$\times$V100 per node, NVLink 300 GB/s, nodes connected via 12.5 GB/s switches). Alpa’s profiling-based variant (Alpa-O), which performs fine-grained intra- and inter-operator sharding, is highly effective at small scales and serves as a strong baseline.
We run Alpa-O on the full cluster, while \nest leverages pre-profiled traces and simulator-guided placement to eliminate extensive profiling overhead. On a scaled-down Mixtral model (Appendix~\ref{app:mixtralmodel}), \nest achieves throughput within 7\% of Alpa on the 8-GPU cluster while reducing optimization time from 1 hour to 5 minutes, and \textbf{outperforms Alpa by 1.8$\times$ on the 16-GPU cluster}. These results show that \nest matches highly optimized, profiling-based strategies on real hardware at small scales while scaling more efficiently to larger clusters.

%% file: body/fatTree_table.tex
\begin{table*}[ht]

\caption{Model workloads evaluated using \nest. Hyperparameters: number of layers (\#L), attention heads (\#AH), and hidden size (H). Distributed strategies for 512 devices specify pipeline depth (p), data-parallel width (d), tensor model-parallel width (t), sequence parallel width (s), expert degree (e), and context-parallel degree (c), formatted as \{p, d, t, s, (e,c)\}. Sequence-parallel width (s), if applied, equals tensor model-parallel width (t), as both are partitioned across the same set of devices. For Alpa, only the number of pipelines per stage is listed, as it performs intra-operator sharding within each stage.}
\vspace{2ex}
\textcolor{black}{
\renewcommand{\arraystretch}{1.4}
\resizebox{\textwidth}{!}{%
\begin{tabular}{ccccccccclccc}
\hline
\multicolumn{1}{c|}{} & \multicolumn{1}{c|}{} & \multicolumn{1}{c|}{} & \multicolumn{1}{c|}{\textbf{Hyper}} & \multicolumn{1}{c|}{} & \multicolumn{1}{c|}{} & \multicolumn{1}{c|}{} & \multicolumn{6}{c}{\textbf{Distributed Strategy \{p,d,t,s,(e,c)\}}} \\ \cline{8-13} 
\multicolumn{1}{c|}{\textbf{Model}} & \multicolumn{1}{c|}{\textbf{Model}} & \multicolumn{1}{c|}{\textbf{Sequence}} & \multicolumn{1}{c|}{\textbf{parameters}} & \multicolumn{1}{c|}{\textbf{TMP}} & \multicolumn{1}{c|}{\textbf{Expert}} & \multicolumn{1}{c|}{\textbf{Context}} & \multicolumn{1}{c|}{\textbf{Manual}} & \multicolumn{1}{c|}{\textbf{MCMC}} & \multicolumn{1}{c|}{\textbf{Alpa}} & \multicolumn{1}{c|}{\textbf{Phaze}} & \multicolumn{1}{c|}{\textbf{Nest}} & \textbf{Recomputation} \\
\multicolumn{1}{c|}{} & \multicolumn{1}{c|}{\textbf{Parameters}} & \multicolumn{1}{c|}{\textbf{Length}} & \multicolumn{1}{c|}{\#L, \#AH, H} & \multicolumn{1}{c|}{\textbf{Widths}} & \multicolumn{1}{c|}{\textbf{Degree}} & \multicolumn{1}{c|}{\textbf{Degree}} & \multicolumn{1}{c|}{} & \multicolumn{1}{c|}{} & \multicolumn{1}{l|}{} & \multicolumn{1}{c|}{} & \multicolumn{1}{c|}{} & \textbf{vs. Stashing} \\ \hline
\multicolumn{1}{c|}{Llama2 7B} & \multicolumn{1}{c|}{7B} & \multicolumn{1}{c|}{4096} & \multicolumn{1}{c|}{32, 32, 4096} & \multicolumn{1}{c|}{-} & \multicolumn{1}{c|}{-} & \multicolumn{1}{c|}{-} & \multicolumn{1}{c|}{\{8, 64, 1, 1\}} & \multicolumn{1}{c|}{\{11, 46, 1, 1\}} & \multicolumn{1}{l|}{\{32, -, -, -\}} & \multicolumn{1}{c|}{\{6, 85, 1, 1\}} & \multicolumn{1}{c|}{\{8, 64, 1, 1\}} & Recomputation \\
\multicolumn{1}{c|}{Llama3 70B} & \multicolumn{1}{c|}{70B} & \multicolumn{1}{c|}{4096} & \multicolumn{1}{c|}{80, 64, 8192} & \multicolumn{1}{c|}{-} & \multicolumn{1}{c|}{-} & \multicolumn{1}{c|}{-} & \multicolumn{1}{c|}{\{80, 6, 1, 1\}} & \multicolumn{1}{c|}{\{80, 6, 1, 1\}} & \multicolumn{1}{l|}{\{52, -, -, -\}} & \multicolumn{1}{c|}{\{41, 12, 1, 1\}} & \multicolumn{1}{c|}{\{81, 6, 1, 1\}} & Recomputation \\ \hline
\multicolumn{2}{c}{\textbf{Tensor Model Parallel Models}} &  &  &  & - & - &  &  &  &  &  &  \\ \hline
\multicolumn{1}{c|}{BertLarge} & \multicolumn{1}{c|}{350M} & \multicolumn{1}{c|}{512} & \multicolumn{1}{c|}{24, 16, 1024} & \multicolumn{1}{c|}{1,2,4,8} & \multicolumn{1}{c|}{-} & \multicolumn{1}{c|}{-} & \multicolumn{1}{c|}{\{8, 64, 1, 1\}} & \multicolumn{1}{c|}{\{2, 256, 1, 1\}} & \multicolumn{1}{l|}{\{24, -, -, -\}} & \multicolumn{1}{c|}{\{1, 512, 1, 1\}} & \multicolumn{1}{c|}{\{1, 512, 1, 1\}} & Stashing \\
\multicolumn{1}{c|}{Megatron GPT3} & \multicolumn{1}{c|}{175B} & \multicolumn{1}{c|}{2048} & \multicolumn{1}{c|}{96, 96, 12288} & \multicolumn{1}{c|}{4,8} & \multicolumn{1}{c|}{-} & \multicolumn{1}{c|}{-} & \multicolumn{1}{c|}{\{32, 4, 4, 1\}} & \multicolumn{1}{c|}{\{9, 7, 8, 1\}} & \multicolumn{1}{l|}{\{72, -, -, -\}} & \multicolumn{1}{c|}{\{16, 8, 4, 1\}} & \multicolumn{1}{c|}{\{16, 8, 4, 4\}} & Recomputation \\ \hline
\multicolumn{2}{c}{\textbf{Expert Parallel Models}} &  &  &  &  &  &  &  &  &  &  &  \\ \hline
\multicolumn{1}{c|}{Mixtral 8x7B} & \multicolumn{1}{c|}{47B} & \multicolumn{1}{c|}{4096} & \multicolumn{1}{c|}{32, 32, 14336} & \multicolumn{1}{c|}{-} & \multicolumn{1}{c|}{1,2,4,8} & \multicolumn{1}{c|}{1,2,4,8} & \multicolumn{1}{c|}{\{32, 4, 1, 1, 4, 1\}} & \multicolumn{1}{c|}{\{32, 4, 1, 1, 4, 1\}} & \multicolumn{1}{l|}{\{32, -, -, -\}} & \multicolumn{1}{c|}{\{16, 8, 1, 1, 4, 1\}} & \multicolumn{1}{c|}{\{8, 8, 1, 1, 4, 2\}} & Recomputation \\ \hline
\end{tabular}
}}
\label{tab:workload}
\vspace{2ex}
\end{table*}

%% file: body/relatedWorks.tex
\section{Related Works}

\paragraph{Device Placement Frameworks.}
Systems such as FlexFlow~\cite{flexflow}, Piper~\cite{piper}, Pip~\cite{pip}, and RL-based approaches~\cite{deviceplacement-rl} automate parallelism configuration using reinforcement learning, dynamic programming, or random search, but typically assume simplified or flat networks and ignore topology heterogeneity. Alpa~\cite{alpa} adds limited topology awareness with a two-level mesh and hybrid ILP–DP search, but lacks support for hierarchical or oversubscribed networks, integrated memory modeling, and expert parallelism.  
Recent work expands the design space along different axes. Aceso~\cite{aesco} emphasizes memory feasibility and utilization via greedy, bottleneck-driven search but relies on local heuristics without global optimality guarantees. Mist~\cite{mist} uses MILP to optimize compute–communication overlap, yet treats topology as secondary and scales poorly to large hierarchical clusters.  
Other systems provide tooling rather than unified optimization. For example, nnScaler~\cite{nnscaler} supports manual exploration of scaling strategies but does not jointly optimize compute, memory, and network placement.

\niparagraph{Joint Optimization with Device Placement.}
Phaze~\cite{phaze} and WHAM~\cite{wham, wham-patent} optimize part of the search space—including hardware and placement—using ILP for operator scheduling but assume a flat network. TopoOpt~\cite{topoopt} adds topology awareness via FlexNet and MCMC-based placement, yet lacks optimality guarantees, scales poorly in large search spaces, and does not support ZeRO or expert parallelism. CATransformers~\cite{catransformers} jointly optimizes model and hardware accelerator configurations, but directly prunes model configurations instead of leveraging parallelization strategies to distribute the model across devices.
\nest advances this direction with a dynamic programming framework for placement on fixed accelerator architectures, unifying communication, memory, and topology modeling to enable hybrid parallelism across hierarchical and oversubscribed networks while preserving scalability and optimality.

\niparagraph{Network-Aware Distributed Machine Learning.}
Works such as Cassini~\cite{cassini} and Themis~\cite{themis} study network-aware scheduling in multi-tenant ML clusters. Cassini uses geometric abstractions for job placement, while Themis coordinates collective operations to reduce congestion. These approaches mitigate inter-job interference, whereas \nest focuses on optimizing a single distributed training job.

%% file: body/conclusion.tex
\section{Limitation and Discussion}


\nest uses a flexible, level-wise abstraction of network locality. While applied here to hierarchical topologies such as fat-tree, it is not limited to them. By decoupling logical locality from physical hierarchy, \nest’s dynamic programming engine generalizes across diverse interconnects.
For non-hierarchical networks (e.g., 3D torus), levels can represent affinity classes based on hop distance or coordinate proximity, with costs derived from profiled or analytical latencies. This enables \nest to adapt to new architectures by updating the cost model without changing the DP formulation.
\nest also provides a structured framework for integrating existing and emerging parallelism, while focusing on placement optimization rather than new parallelization techniques.




\section{Conclusion}
We presented \nest, the first structured, network-, compute-, and memory-aware device placement framework that unifies model parallelism and placement via structured dynamic programming. By modeling hierarchical networks and supporting a wide variety of parallelism strategies, \nest enables efficient, scalable training across diverse hardware. Evaluations demonstrate consistent gains in throughput, memory efficiency, and scalability over state-of-the-art baselines, providing a foundation for co-designing parallelization strategies and AI datacenter infrastructure.

\section{Acknowledgments}
We thank Amar Phanishayee for discussions that helped inspire and shape the initial direction of this project, as well as our shepherd and reviewers for their insightful comments. This project is partially supported by gifts from Google, AMD, and Natural Sciences and Engineering Research Council of Canada (NSERC) [funding reference number 587440-2024]. This research was supported through cyber-infrastructure research resources and services provided by the Partnership for an Advanced Computing Environment (PACE) at the Georgia Institute of Technology. This work was also supported in part by ACE, one of the seven centers in JUMP 2.0, a Semiconductor Research Corporation (SRC) program sponsored by DARPA.

%% file: body/appendix.tex
\section*{Appendix}
\appendix

\section{Artifact Appendix}

\subsection{Abstract}
\nest is a network-, compute-, and memory-aware framework for automatic device placement that unifies model parallelism and placement via structured dynamic programming for large-scale distributed training. This artifact provides the source code, bash scripts, and instructions necessary to reproduce the key results in Section~\ref{sec:eval} for \nest and all evaluated baselines. \nest is compatible with any NVIDIA GPU supporting PyTorch 2.5 and CUDA 12.4 with at least 100\,GB of RAM. When using the provided operator graphs and estimates included as part of this artifact, most experiments can also be executed in CPU-only environments with at least 64\,GB of RAM. Note that an active Gurobi license is required to run the \nest solver; however, most academic users can obtain a free Gurobi WLS license.

\subsection{Artifact check-list (meta-information)}

{\small
\begin{itemize}
  \item {\bf Algorithm:} Dynamic programming-based network-aware automatic device placement. 
  \item {\bf Program:} C++ solver for the DP algorithm; Python for graph extraction and workflow management.
  \item {\bf Compilation:} C++ solver compiled with \texttt{g++} version 12.4.0.
  \item {\bf Data set:} Large language model operator graphs (GPT-3 175B, Llama2-7B, Llama3-70B, BertLarge, Mixtral-8x7B) from HuggingFace Transformers and Megatron-LM (included in the artifact).
  \item {\bf Run-time environment:}
  \begin{itemize}
      \item Python 3.10, PyTorch 2.5 with \texttt{torch.fx}, and Gurobi Optimizer.
      \item Anaconda/Miniconda for environment management.
      \item Linux, primarily tested on RHEL 9.6.
      \item Note: An active Gurobi license is required to run the solver (most academic users can obtain a free Gurobi WLS license).
      \item Collecting operator graphs from scratch requires CUDA 12.4 or similar.
  \end{itemize}
  \item {\bf Hardware:}
  \begin{itemize}
      \item Ideal: Any GPU supporting PyTorch 2.5 and CUDA 12.4 with at least 100\,GB RAM.
      \item If using provided operator graphs/estimates: any CPU with 64\,GB RAM is sufficient for most experiments.
      \item Profiling operator latency from scratch requires an NVIDIA H100 GPU.
  \end{itemize}
  \item {\bf Execution:} A single GPU or CPU is sufficient for execution.
  \item {\bf Metrics:} Training throughput, optimization time, and scalability up to 1,024 GPUs/accelerators.
  \item {\bf Output:} Throughput improvement for each baseline is printed to the console and saved as a plot. Raw throughput results are saved in a .csv file.  
  \item {\bf Experiments:} Prepared with bash scripts and detailed instructions in the README. Throughput results should be deterministic; however, runtime results may vary based on the execution environment (reported runtime results were primarily obtained on H100 GPUs).
  
  \item {\bf Disk space required:} Under 30\,GB for source code, profiles, model graphs, and the Conda environment.
  \item {\bf Preparation time:} Approximately 1 hour.
  \item {\bf Experiment time:} Approximately 4 hours using the provided profiles and model graphs; over 5 days to collect Alpa-E baseline results (excluded from the Artifact Evaluation).
  \item {\bf Publicly available:} Yes.
  \item {\bf Code licenses:} MIT License.
  \item {\bf Archived (DOI):} 10.5281/zenodo.18826203
\end{itemize}
}

\subsection{Description}

\subsubsection{How delivered}
The source code and scripts are available at \href{https://github.com/scai-tech/Nest}{https://github.com/scai-tech/Nest}. 

The artifact is also publicly available as a Trovi artifact at  \href{https://trovi.chameleoncloud.org/artifacts/a8a96d3d-4921-448e-a9fa-09db5950e26d/}{https://trovi.chameleoncloud.org/artifacts/a8a96d3d-4921-448e-a9fa-09db5950e26d/}

\subsubsection{Hardware dependencies}
Experiments were conducted on H100 and V100 GPUs and also tested on RTX 6000 GPUs (CUDA 12.4, at least 100\,GB RAM). When using the provided operator graphs and estimates, any GPU or CPU supporting PyTorch 2.5 or higher with at least 64\,GB RAM is sufficient for most experiments.

\subsubsection{Software dependencies}
\begin{itemize}
    \item Requires a Linux environment (tested with RHEL 9.6 and Ubuntu 22.04) with Anaconda or Miniconda. All Python dependencies, including PyTorch 2.5 and the Gurobi Python interface, are managed via the provided \texttt{environment.yml} file.
    \item \texttt{g++} version 12.4.0 or higher for the DP solver.
    \item Gurobi Optimizer license (academic users can obtain a free Gurobi WLS license).
    \item Python 3.10 with PyTorch 2.5.
    \item Astra-Sim and Sunstone simulators (included with the source code).
\end{itemize}

\subsubsection{Data sets and Models}
Extracted operator graphs for all evaluated LLM models are included in the GitHub repository.

\subsection{Installation}

\begin{lstlisting}[breaklines=true, basicstyle=\ttfamily\fontsize{8pt}{7.8pt}\selectfont]
git clone https://github.com/scai-tech/Nest.git
cd Nest
conda env create -f environment.yml
conda activate nestenv
\end{lstlisting}
By default, \texttt{conda activate} should set \texttt{\$CONDA\_PREFIX} automatically. Verify this by running:
\begin{lstlisting}[breaklines=true, basicstyle=\ttfamily\fontsize{8pt}{7.8pt}\selectfont]
echo $CONDA_PREFIX
\end{lstlisting}
If it is not set, please manually export it to point to your Conda environment directory. Then, build the DP solver and simulator components:
\begin{lstlisting}[breaklines=true, basicstyle=\ttfamily\fontsize{8pt}{7.8pt}\selectfont]
./setup.sh
\end{lstlisting}
\begin{lstlisting}[breaklines=true, basicstyle=\ttfamily\fontsize{8pt}{7.8pt}\selectfont]
source $CONDA_PREFIX/etc/conda/activate.d/nest_paths.sh
\end{lstlisting}
\textbf{[Not required for AE]} If running on a GPU supporting CUDA 12.4 and you wish to extract graphs from scratch, install the APEX library:
\begin{lstlisting}[breaklines=true, basicstyle=\ttfamily\fontsize{8pt}{7.8pt}\selectfont]
./setup.sh --apex_only
\end{lstlisting}

\subsection{Experiment workflow}
We provide scripts for reproducing all results. \textbf{We recommend following the \texttt{README.md}, which provides detailed explanations for each step.}

\begin{enumerate}
    \item (Optional) Run scripts to collect operator graphs and estimates/profiles from scratch.
    \item (Optional) Collect Alpa baseline results.
    \item Run the evaluation scripts to execute \nest and the compared baselines (Manual, MCMC, Phaze) across the models and network configurations evaluated in the paper.
\end{enumerate}

Note: Alpa-E experiments have specific hardware requirements and a long execution time; they are therefore excluded from the standard artifact evaluation. Instead, we provide the pre-collected results presented in the paper. Full instructions for running Alpa-E are available in the repository for completeness.

\subsection{Evaluation and expected results}
The artifact reproduces the results shown in Figures~\ref{fig:throughput} and~\ref{fig:batch} and Table~\ref{fig:h100}. Outputs are printed to the terminal and saved to \texttt{scripts/<setup\_name>/out}; plots are saved to \texttt{scripts/<setup\_name>/plots}. The \texttt{<setup\_name>} values are \texttt{tpuv4\_fatTree} and \texttt{h100\_spineLeaf}, corresponding to the network setups evaluated in the paper.

Expected overall runtimes are based on H100 GPU execution on an HPC cluster with network-mounted storage. Local machines with SSD storage may run faster. First-time runs may be slower due to Python and CUDA compilation caching. Actual \nest solving times are also reported in the execution logs.

\begin{enumerate}
    \item \textbf{Figure~\ref{fig:throughput} (TPUv4, Llama2-7B):} Reproduce results for the 7B model [$\approx$10 minutes].
    \item \textbf{Figure~\ref{fig:throughput} (All models):} Reproduce results for BertLarge, Llama2-7B, Llama3-70B, GPT-3 175B, and Mixtral-8x7B [$\approx$2 hours].
    \item \textbf{Figure~\ref{fig:batch} (Batch size sweep):} Reproduce the micro-batch size sweep with Llama2-7B [$\approx$20 minutes].
    \item \textbf{Figure~\ref{fig:h100} (H100, Mixtral):} Reproduce spine-leaf topology results for Mixtral [$\approx$10 minutes].
\end{enumerate}

\subsection{Experiment customization}
The \texttt{README.md} provides instructions for running \nest with custom network configurations, models, and other parameters.


\section{Network Topology}
\subsection{Hierarchical Network Topology}
\label{app:topology}

\begin{figure}[htbp]
    \centering
    \begin{subfigure}[b]{0.5\textwidth}
        \centering
        \includegraphics[trim=0 11cm 0 2cm, clip, width=1\linewidth]{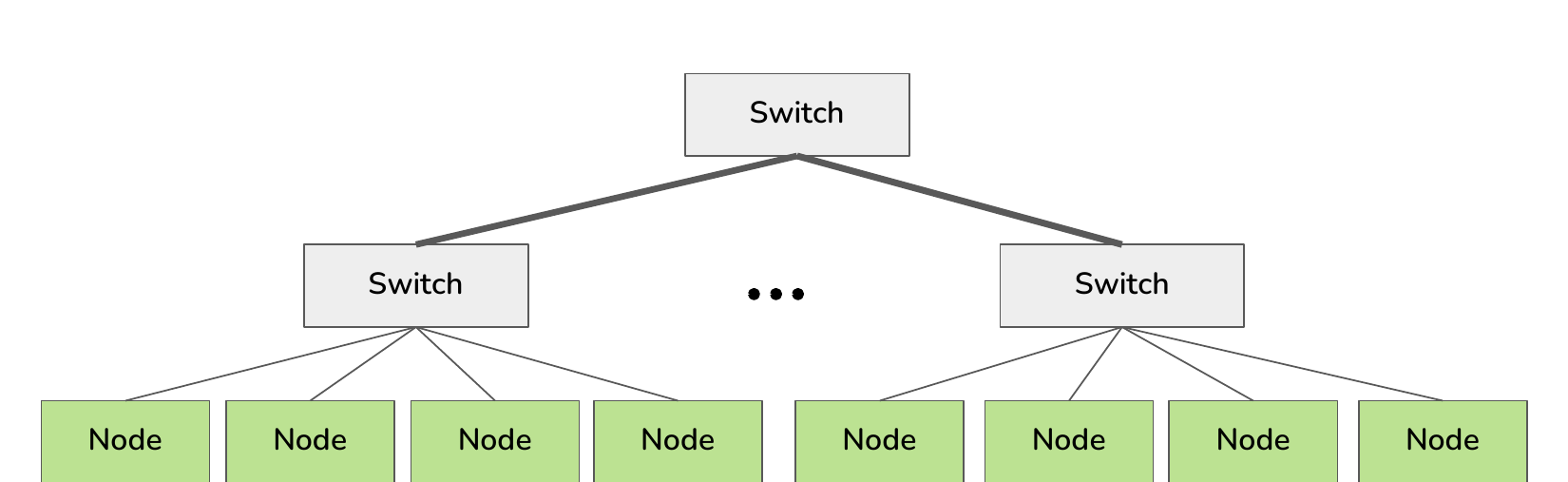}
        \caption{Multi-level Hierarchical Tree Topology}
        \label{fig:tree}
    \end{subfigure}
    \hfill
    \begin{subfigure}[b]{0.5\textwidth}
        \centering
        \includegraphics[trim=0 11cm 0 2cm, clip, width=1\linewidth]{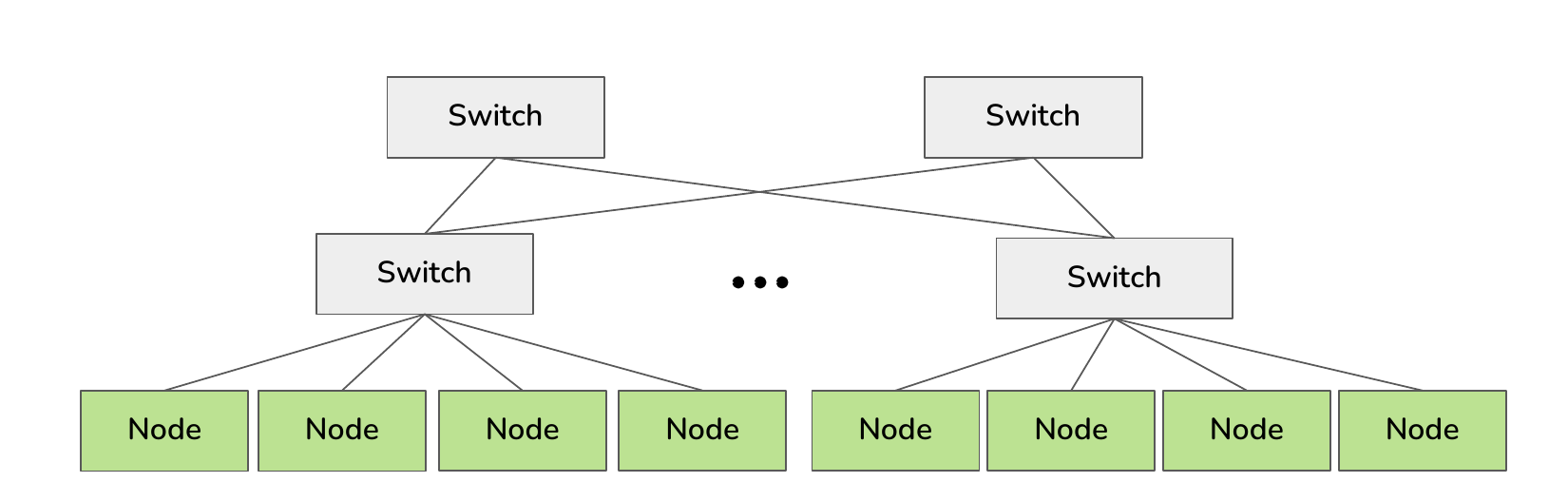}
        \caption{Spine-Leaf Topology}
        \label{fig:spine}
    \end{subfigure}
    \hfill
    \begin{subfigure}[b]{0.5\textwidth}
        \centering
        \includegraphics[trim=0 11cm 0 2cm, clip, width=1\linewidth]{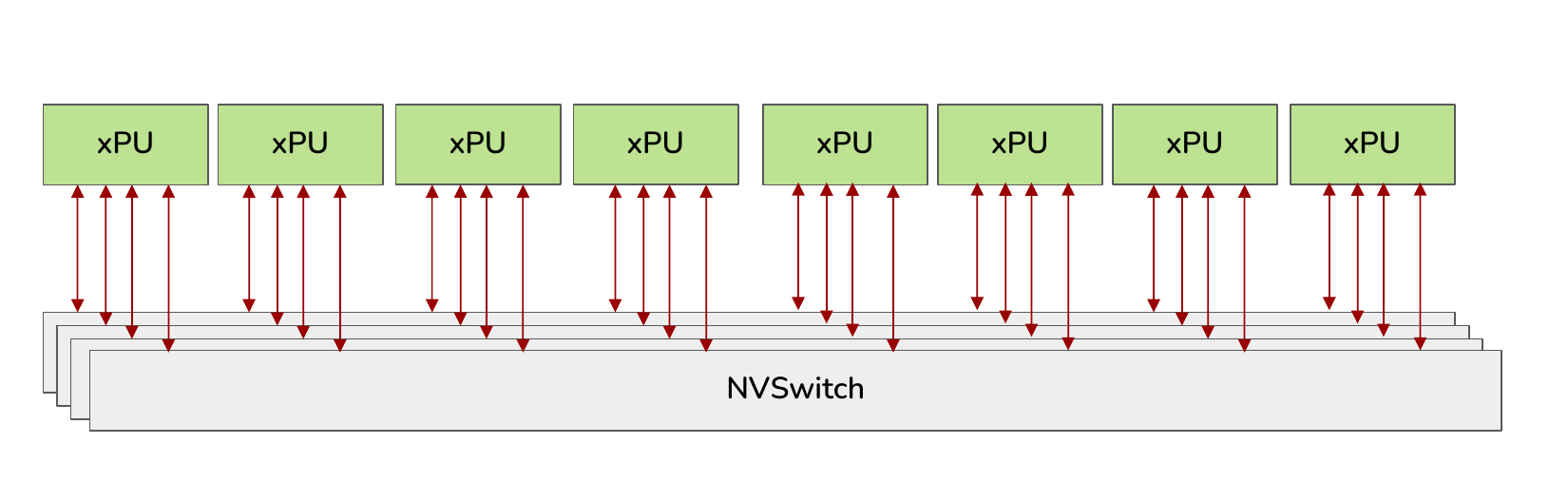}
        \caption{Intra-Node, HGX-style~\cite{hgx}}
        \label{fig:hgx}
    \end{subfigure}
    \vspace{-2ex}
    \caption{Hierarchical network topology. Black links represent InfiniBand connections, while red links represent NVLink connections.}
    \label{fig:network}
\end{figure}

Modern data centers commonly adopt multi-level hierarchical network topologies such as Spine–Leaf or Clos networks, which provide high aggregate bandwidth while maintaining modularity and fault tolerance. While the Fat-Tree topology is often used as a simplified abstraction in evaluation settings, real-world clusters frequently implement oversubscribed or non-uniform variants of hierarchical networks due to cost, physical layout constraints, or workload demands. Figure~\ref{fig:network} illustrates representative examples of these topologies.

\nest supports a wide range of such configurations through a flexible network modeling interface. Users provide a network description specifying device identifiers, node connectivity, per-link bandwidth and latency, and the collective communication protocols used by the system. This abstraction allows \nest to model both idealized and real-world hierarchical deployments without requiring topology-specific modifications.

\subsection{Mesh and Torus Modeling}
\label{app:network_mesh}

For completeness, we also describe how the framework can be extended to non-hierarchical topologies such as meshes and tori (e.g., TPU clusters). In these architectures, communication cost is determined primarily by physical distance between devices. \nest captures this structure by mapping physical proximity to logical communication levels.

\begin{figure}[H]
\vspace{2ex}
\centering
\includegraphics[width=0.6\columnwidth]{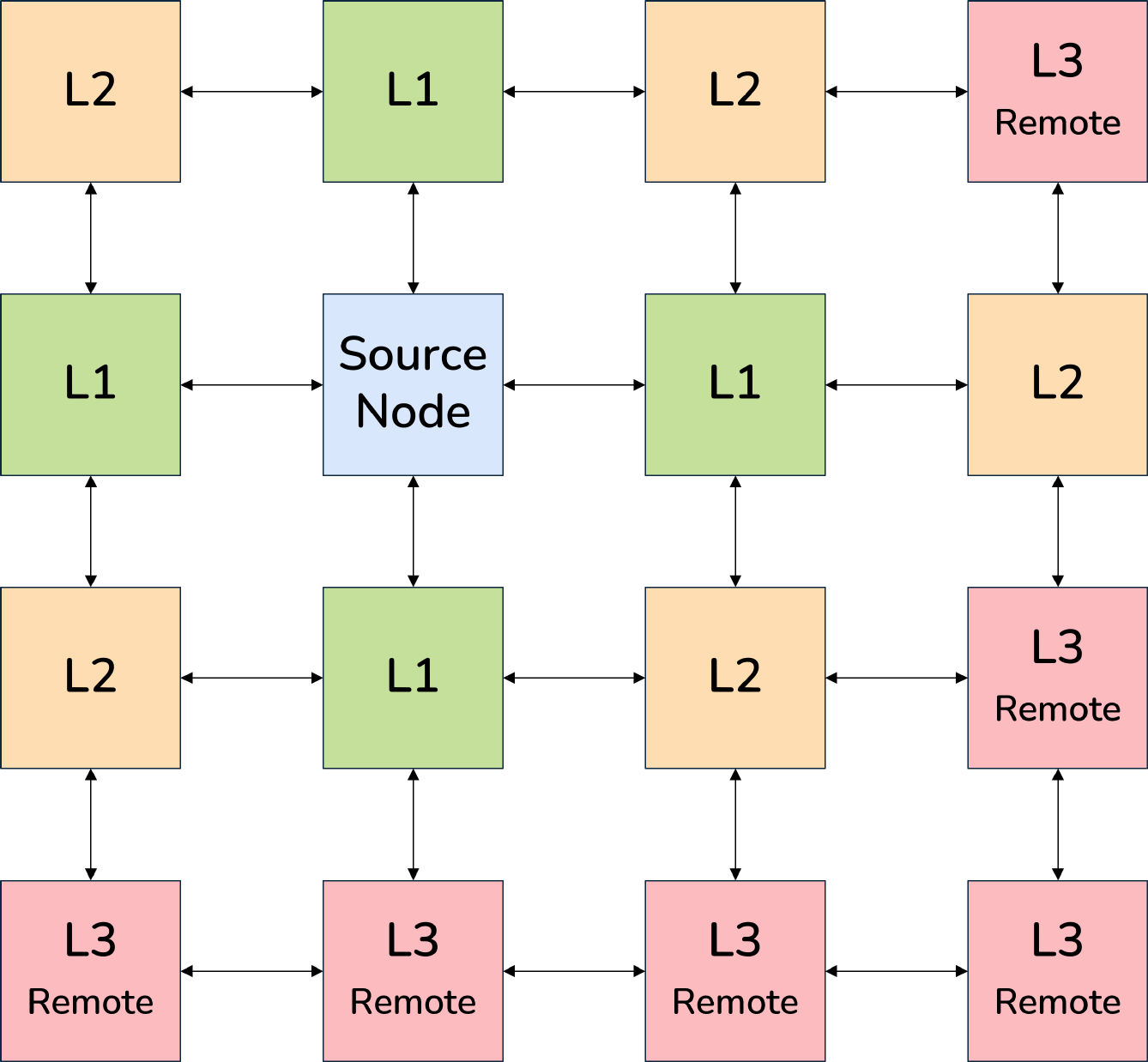}

\caption{Level-Wise Abstraction for Mesh/Torus Topologies. \nest maps physical hop distances to discrete communication levels ($l$) relative to a reference source node. This abstraction preserves optimal substructure in the DP by grouping nodes into affinity classes, where each class corresponds to a modeled communication latency or bandwidth constraint.}
\label{fig:network_mesh}
\end{figure}

As illustrated in Figure~\ref{fig:network_mesh}, \nest defines coordinate-based affinity classes based on Manhattan hop distance or heterogeneous link bandwidth relative to a source node:

\begin{itemize}[leftmargin=*, itemsep=1pt, parsep=0pt, topsep=1pt, partopsep=1pt]
\item \textbf{Level 1 ($L1$):} Immediate neighbors (1-hop) or high-bandwidth local tiles with the lowest communication latency.
\item \textbf{Level 2 ($L2$):} Nodes at a two-hop distance or connected through intermediate-bandwidth links.
\item \textbf{Level 3 ($L3$ / Remote):} Nodes beyond a predefined distance threshold or those connected through lower-bandwidth inter-mesh links.
\end{itemize}

This mapping enables the dynamic programming solver to reason about mesh distance and heterogeneous link bandwidth using the same abstraction used for hierarchical switch tiers. By grouping devices according to communication performance characteristics, the placement engine remains topology-agnostic while still capturing the latency and bandwidth constraints imposed by the physical network.

\section{Extended Evaluation}

\subsection{H100 Spine-Leaf Evaluation}

\subsubsection{Model Configuration for GPT3-35B}
\label{app:gptsmall}
Table~\ref{tab:gpt-config} shows the scaled-down GPT-3 model (GPT3-35B) used in Section~\ref{sec:spine_large}. Reduced from 175B parameters to allow comparison with Mist, which supports hidden dimensions $<$8192, contrast with GPT3-175B’s original hidden size of 12,288. A microbatch size of 1 was used in all experiments.

\begin{table}[H]
\centering
\caption{Configuration of the scaled-down GPT3 model.}
\vspace{2ex}
\begin{tabular}{l r}
\hline
\textbf{Parameter} & \textbf{Value} \\
\hline
Number of Layers & 64 \\
Hidden Dimension & 8192 \\
Number of Heads & 64 \\
Intermediate Dimension & 16384 \\
Sequence Length & 2048 \\
\hline
\end{tabular}
\label{tab:gpt-config}
\end{table}

\subsubsection{Runtime Comparison with Mist}
\label{app:runtime}

\nest on average achieves over 30\% faster runtime than Mist across all evaluated models. 
Table~\ref{tab:runtime} presents the breakdown of each model runtime for each framework.
\vspace{-2ex}
\begin{table}[h]
\centering
\caption{Runtime comparison of \nest against baseline.}
\resizebox{\columnwidth}{!}{%
\begin{tabular}{lccc}
\hline
\textbf{Model} & \textbf{Baseline (min)} & \textbf{\nest (min)} & \textbf{Time Reduction (\%)} \\
\hline
GPT-3 35B    & 17 & 15  & 11.8\% \\
Llama3 70B   & 30 & 6   & 80.0\% \\
Llama2 7B    & 8  & 2.3 & 71.3\% \\
BertLarge    & 3  & 2   & 33.3\% \\
\hline
\end{tabular}
}
\label{tab:runtime}
\end{table}

\subsection{System Validation}

\subsubsection{Model Configuration for Scaled-down Mixtral}
\label{app:mixtralmodel}
Table~\ref{tab:model-config} shows the configuration of the scaled-down Mixtral 8×7B model used in Section~\ref{sec:spine_alpa}. This model was reduced from the full 47B parameter version to enable feasible profiling and end-to-end execution on the resource-constrained 8- and 16-GPU V100 validation clusters. This model has a total of 790M parameters. Microbatch size of 1 used for the experiments. 

\begin{table}[H]
\centering
\caption{Configuration of the scaled-down Mixtral model.}
\label{tab:model-config}
\vspace{2ex}
\begin{tabular}{l r}
\hline
\textbf{Parameter} & \textbf{Value} \\
\hline
Number of Layers & 8 \\
Number of Experts & 8 \\
Hidden Dimension & 1024 \\
Number of Heads & 16 \\
Intermediate Dimension & 3584 \\
Sequence Length & 1024 \\
\hline
\end{tabular}
\end{table}

\subsubsection{Memory Estimation}
\label{app:memory}

\nest memory per-layer modeling estimates are, on average,  within 7\% of compiled executables from Alpa~\cite{alpa} across evaluated models.
\vspace{-2ex}
\begin{table}[h]
\centering
\caption{Per layer Memory requirement estimations between profiled Alpa executables and \nest estimations}
\resizebox{\columnwidth}{!}{%
\begin{tabular}{lccc}
\hline
\textbf{Model} & \textbf{Alpa Executables (GB)} & \textbf{\nest Estimates (GB)} \\
\hline
GPT-3 175B    & 10.1 & 9.7 \\
Llama3 70B   & 24.8 & 22.9   \\
Llama2 7B    & 9.8  & 8.1  \\
BertLarge    & 0.21 & 0.21   \\
\hline
\end{tabular}
}
\label{tab:memory}
\end{table}

\vspace{-2ex}

\subsubsection{Hardware Validation}
\label{app:val}

\begin{figure}[h]
\centering
\includegraphics[width=0.95\columnwidth]{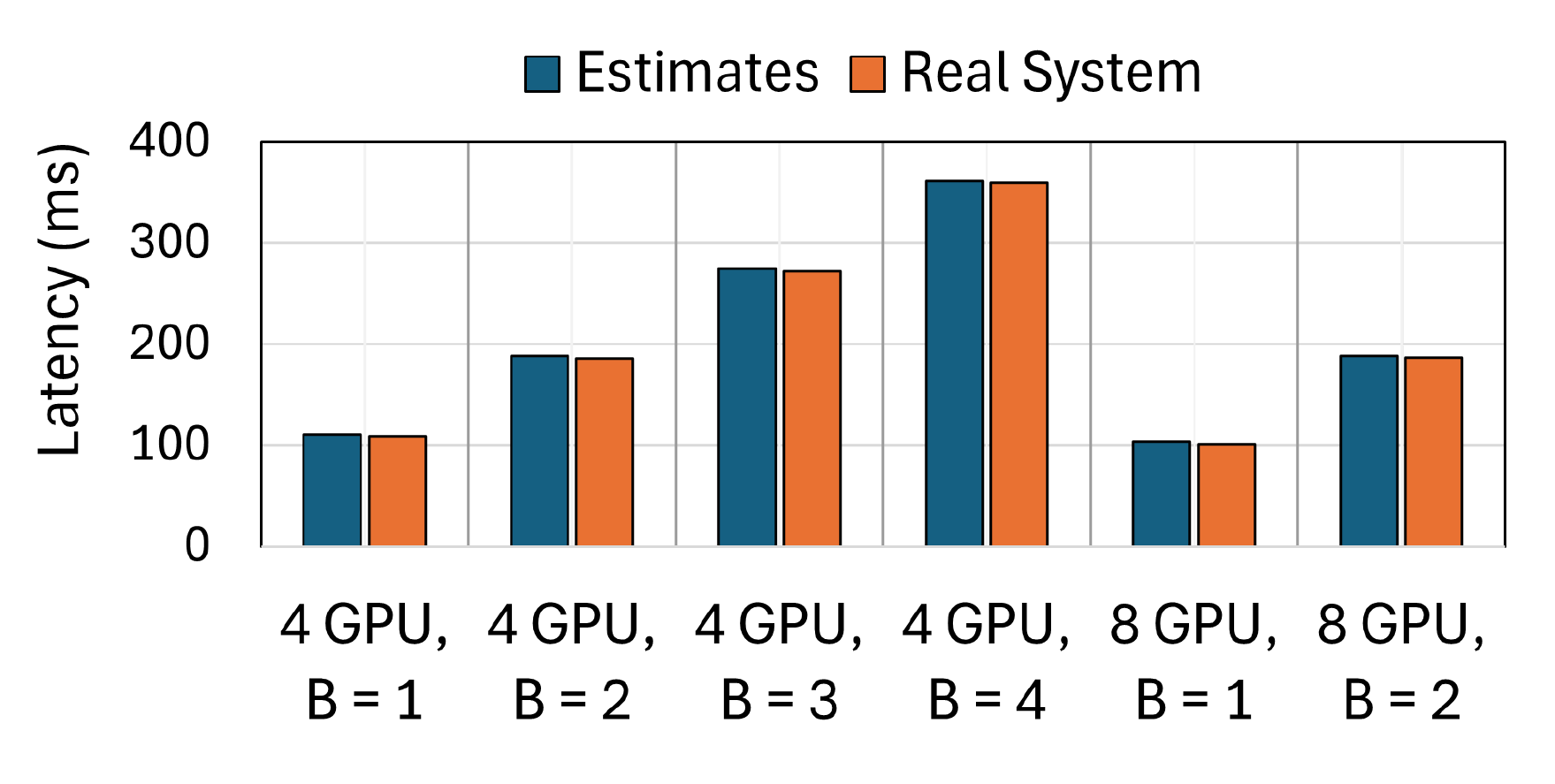}
\vspace{-2ex}
\caption{Validation of Collective Communication Estimations~\cite{astrasim} against H100 GPU nodes}
\label{fig:validation}
\vspace{-1ex}
\end{figure}

We validated our framework’s collective communication estimates against real-system measurements on clusters of 4 and 8 NVIDIA H100 SXM5 80GB GPUs~\cite{h100-nvidia}. The GPUs are connected via a switch-based topology with four NVSwitches, each linking eight GPUs through NVLinks offering 450GB/s unidirectional bandwidth. Figure~\ref{fig:validation} compares end-to-end iteration time (forward+backward) for GPT-3 across batch sizes 1–4, showing up to 2\% latency difference between measured and predicted results. Computation latency was measured using PyTorch, while communication latency was simulated.
\vspace{-1ex}

\subsection{ZeRO Optimization Ablation Study}
\label{app:zero}

\nest incorporates memory optimization techniques, including ZeRO and activation recomputation, to overcome memory bottlenecks and enable training configurations that would otherwise be infeasible. ZeRO is most beneficial when even a single model layer exceeds device memory. Using detailed memory modeling, \nest automatically detects such constraints and applies the appropriate ZeRO stage (1, 2, or 3) as needed, generating valid and efficient training strategies under tight memory budgets.

Although most evaluated models fit within current hardware (TPUv4 64~GB HBM, H100 80~GB HBM), we perform ablations with reduced-memory configurations to isolate ZeRO’s impact. As shown in Table~\ref{tab:zero}, \nest assigns each layer—including embeddings—to a single pipeline stage and selectively applies ZeRO partitioning, confirming that training becomes infeasible without it.
For each model, \nest determines the optimal ZeRO configuration at the layer level, identifying which states (\emph{optimizer}, \emph{gradient}, or \emph{parameter}) to partition and how many devices to allocate. This fine-grained strategy ensures every layer fits in memory while minimizing cross-device communication.
\input{body/zero_table}

\subsection{Microbatch size Scaling}
\label{app:batch}

\begin{figure*}[ht]
\centering
\includegraphics[width=2\columnwidth]{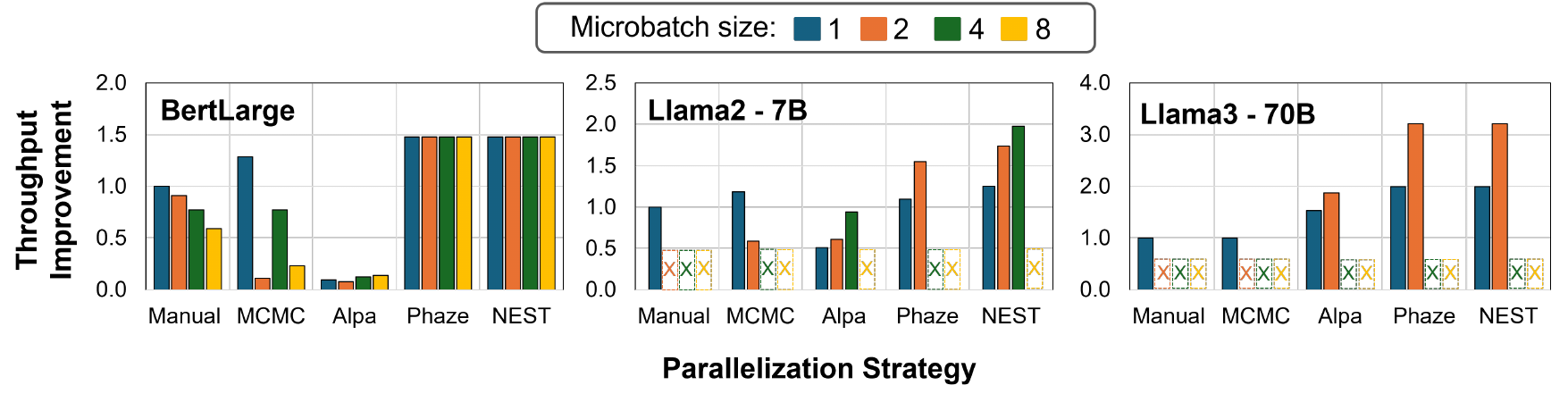}
\vspace{-2ex}
\caption{Throughput improvement relative to manual strategy with microbatch size 1 across different parallelization strategies. “X” indicates cases where the baseline fails to find a feasible placement within memory constraints. Llama2-7B and Llama3-70B are realizable only up to batch size 4 and 2 respectively due to high memory requirements.}
\label{fig:batch_512}
\vspace{-3ex}
\end{figure*}

Figure~\ref{fig:batch_512} shows microbatch performance for BertLarge, Llama2-7B, and Llama3-70B on a 512-device cluster. Trends mirror those at 256 devices (Section 5.2), with optimal microbatch size depending on model and parallelization. Alpa benefits from larger microbatches, improving up to 1.5× from size 1 to 8, but its performance on BertLarge drops at 512 devices due to excessive oversharding.

\section{Algorithm for device placement}
\label{app:dpalg}
Algorithm~\ref{alg:full} walks through the full pseudo code of the network-aware device placement algorithm.
\vspace{1ex}  

\begin{figure*}[t]
\centering
\noindent
\begin{minipage}{0.95\textwidth}
\begin{algorithm}[H]
\caption{Device Placement Optimization with Latency and Memory Considerations}
\label{alg:full}
\begin{algorithmic}[1]
\REQUIRE Layerwise latency/memory estimates, architecture \& network configs, SUB-GRAPH config $sg$
\ENSURE Device placement, $t_{\text{batch}}$
\STATE \textbf{Init:} $dp[\ell][id][k][s] \gets (\infty, \emptyset)$ \algcomment{$dp$[level][downset][accelerators][stages]}
\STATE $dp[\ell][\emptyset][*][0] \gets (0, \text{init})$ \algcomment{Base case: no layers assigned}
\FOR{$id \in \mathit{Downsets} \setminus \{D_{\text{full}}\}$}
    \STATE $\mathit{isZeRO} \gets \textsc{IsZeroMemory}(\textsc{LayerMemReq}(id))$ \algcomment{Apply ZeRO if needed}
    \FOR{$\mathit{subId} \subsetneq id,\ \mathit{load} \gets id \setminus \mathit{subId}$}
        \STATE $\mathit{loads} \gets \textsc{GetLoadOfStage}(\mathit{load}, \mathit{isZeRO})$ \algcomment{Latencies for all placements}
        \FOR{$s \in [1, S_{\max}],\ k \in [1, K_{\max}],\ a \in \textsc{ValidAccelCounts}(\mathit{load})$}
            \FOR{each pair of levels $\ell, \ell'$}
                \STATE $\mathit{prev} \gets dp[\ell'][\mathit{subId}][k{-}a][s{-}1]$
                \STATE $c \gets \max(\mathit{prev}.\text{latency},\ \mathit{loads}[\ell][a].\text{latency})$
                \IF{$c < dp[\ell][id][k][s].\text{latency}$}
                    \STATE $dp[\ell][id][k][s] \gets (c,\ (\mathit{subId}, a))$ \algcomment{Update min--max latency and backpointer}
                \ENDIF
            \ENDFOR
        \ENDFOR
    \ENDFOR
\ENDFOR
\STATE $t_{\min} \gets \infty$ \algcomment{Find best end-to-end batch time}
\FOR{$\mathit{subId} \subsetneq D_{\text{full}},\ F \gets D_{\text{full}} \setminus \mathit{subId}$}
    \STATE $\mathit{load} \gets \textsc{GetLoadOfStage}(F, \mathit{isZeRO})$ \algcomment{Latency for first stage}
    \FOR{$s \in [1, S_{\max}],\ k \in [1, K_{\max}],\ a \in \textsc{ValidAccelCounts}(F)$}
        \FOR{each pair of levels $\ell, \ell'$}
            \STATE $\mathit{prev} \gets dp[\ell'][\mathit{subId}][k{-}a][s{-}1]$
            \STATE $t_{\text{stage}} \gets \max(\mathit{prev}.\text{latency},\ \mathit{load}[\ell][a].\text{latency})$
            \STATE $t_{\text{batch}} \gets t_{\text{stage}} \cdot (mbs + s - 1) + \textsc{SyncCost}(F)$ \algcomment{Pipeline + sync}
            \IF{$t_{\text{batch}} < t_{\min}$}
                \STATE $t_{\min} \gets t_{\text{batch}}$
            \ENDIF
        \ENDFOR
    \ENDFOR
\ENDFOR
\STATE \textbf{return} $\textsc{ReconstructStages}(dp,\ D_{\text{full}})$ \algcomment{Retrieve final device placement}
\end{algorithmic}
\end{algorithm}
\end{minipage}
\end{figure*}

%% file: body/zero_table.tex
\begin{table}[h]
\vspace{-2ex}
\caption{Models evaluated with resource-constrained architectures. Distributed strategy formatted as \{p,d,t\}}
\renewcommand{\arraystretch}{1.25} 
\resizebox{\columnwidth}{!}{%
\begin{tabular}{cc|c|c}
\hline
\multicolumn{2}{c|}{\textbf{Model}} & \textbf{Llama3 70B} & \textbf{BertLarge} \\ \hline
\multicolumn{2}{c|}{\textbf{HBM}} & 24GB &  120MB \\ \hline
\multicolumn{2}{c|}{\textbf{Number of Device Used}} & 641 & 980 \\ \hline
\multicolumn{2}{c|}{\textbf{\begin{tabular}[c]{@{}c@{}}Distributed Strategy\\\end{tabular}}} & \{81, 1, 1\} &  \{25, 10, 1\}\\ \hline
\multicolumn{1}{c|}{\multirow{2}{*}{\textbf{ZeRO}}} & \textbf{Layer 0 (embedding)} & None & ZeRO-2: (degree 2) \\
\multicolumn{1}{c|}{} & \textbf{Remaining Layers} & ZeRO-3: (degree 8) & ZeRO-3: (degree 4) \\ \hline
\end{tabular}%
}
\vspace{-3ex}
\label{tab:zero}
\end{table}